\documentclass{article}

\usepackage[preprint]{neurips_data_2023}
\usepackage[utf8]{inputenc} % allow utf-8 input
\usepackage[T1]{fontenc}    % use 8-bit T1 fonts
\usepackage{hyperref}       % hyperlinks
\usepackage{url}            % simple URL typesetting
\usepackage{booktabs}       % professional-quality tables
\usepackage{amsfonts}       % blackboard math symbols
\usepackage{nicefrac}       % compact symbols for 1/2, etc.
\usepackage{microtype}      % microtypography
\usepackage{graphicx}
\usepackage{amsmath}
\usepackage{amssymb}
\usepackage{algorithm}
\usepackage{algpseudocode}
\usepackage{subcaption}
\usepackage{lipsum}
\usepackage{wrapfig}
\usepackage{makecell}
\usepackage[table]{xcolor}

\title{Assessing Uncertainty Estimation Methods for 3D Image Segmentation under Distribution Shifts}
\author{%
  Masoumeh Javanbakhat \\
  Digital Health-Machine Learning\\
  Hasso-Plattner-Institute\\
  Potsdam, 14482, Germany \\
  \texttt{Masoumeh.Javanbakhat@hpi.de} \\
 \And
 Md Tasnimul Hasan \\
 Faculty of Sciences \\
 University of Potsdam \\
 Potsdam, 14482, Germany \\
 \texttt{tasnimul.abir@gmail.com} \\
 \And
 Cristoph Lippert \\
 Digital Health-Machine Learning \\
 Hasso-Plattner-Institute  \\
 Potsdam, 14482, Germany \\
 \texttt{Christoph.Lippert@hpi.de} \\
}

\begin{document}
\bibliographystyle{abbrvnat}
\maketitle

\begin{abstract}
%In recent years, machine learning approaches have witnessed extensive adoption across various sectors, ranging from spam detection to chatGPT. However, computer-assisted image-based disease detection and diagnosis, leveraging deep learning methods, remain uncommon practices among medical professionals and practitioners. One of the primary reasons for this is that typical deep learning approaches assume the use of training and test datasets with identical distributions. 
In recent years, machine learning has witnessed extensive adoption across various sectors, yet its application in medical image-based disease detection and diagnosis remains challenging due to distribution shifts in real-world data. In practical settings, deployed models encounter samples that differ significantly from the training dataset, especially in the health domain, leading to potential performance issues.
%This limitation renders the deep learning models less expressive and less reliable to the experts in the health domain. 
This limitation hinders the expressiveness and reliability of deep learning models in health applications. Thus, it becomes crucial to identify methods capable of producing reliable uncertainty estimation in the context of distribution shifts in the health sector. In this paper, we explore the feasibility of using cutting-edge Bayesian and non-Bayesian methods to detect distributionally shifted samples, aiming to achieve reliable and trustworthy diagnostic predictions in segmentation task. Specifically, we compare three distinct uncertainty estimation methods, each designed to capture either unimodal or multimodal aspects in the posterior distribution. Our findings demonstrate that methods capable of %capturing \textit{diverse modes} in the posterior distribution,particularly those 
addressing \textit{multimodal} characteristics in the posterior distribution, offer more dependable uncertainty estimates. 
%even in the presence of minor distribution shifts.
This research contributes to enhancing the utility of deep learning in healthcare, making diagnostic predictions more robust and trustworthy.
\end{abstract}

\section{Introduction}\label{sec:intro}
%One paragraph: Problems regarding shift of distribution of data in medical domain
%Second paragraph: UQ methods to detremine OOD samples 
%In the rest of this paragraph, I should again highlight the problem of dist shift

Deep neural networks (DNNs) are known for their promising results in a variety of domains. As a result, the predictive distributions of these models are increasingly being used to make decisions in important applications ranging from machine-learning aided medical diagnoses to self-driving cars. The self-deployment of DNNs in such high-risk scenarios necessitates not only point predictions but also accurate quantification of predictive uncertainty.

Typical DNNs perform under the assumption that the training and test datasets share the same underlying distribution. However, this assumption becomes a notable limitation when the deployed model encounters samples from a different setting. In such scenarios, having calibrated predictive uncertainty becomes crucial. Accurate calibrated uncertainty offers valuable insights, aiding in a precise evaluation of risks. This, in turn, empowers practitioners to grasp potential accuracy fluctuations and grants the system the ability to refrain from decision-making in instances of low confidence.
%Calibrated uncertainty provides valuable insights by facilitating an accurate assessment of risk, enabling practitioners to understand how accuracy might degrade, and allowing the system to abstain from making decisions when confidence is low.

One of the settings that distribution shift really matters is medical image analysis, where the shifts such as: orientation invariance (observed, for example, in cell images), high variance in feature scale (common in X-ray images), and locale specific features (e.g. CT) are prevalent. Some other shifts in medical domain include for example different scanners. A model trained on a dataset collected with a scanner might not be able to interpret sound of the images collected from a different scanner. In the medical diagnostic and screening realm, this issue takes on utmost importance, as errors can lead to profound consequences, making it exceptionally critical to address effectively.

A variety of methods are applied for quantifying predictive uncertainty in DNNs. Inherently, Bayesian neural networks (BNNs) offer the property of uncertainty estimation that results from the diversity of learned representations. They learn a posterior distribution over the parameters that quantify parameter uncertainty, known as \textit{epistemic uncertainty}. There are some non-Bayesian approaches like ensemble of deterministic models to quantify uncertainties.
%The posterior distribution is a complex and multimodal distribution \citep{cSGMCMC}.
Among the proposed uncertainty estimation methods, some of them focus on taking samples from one mode, while others possess the capability to capture diverse modes in the posterior distribution. It is noteworthy that the posterior distribution is a complex and multimodal distribution \citep{bishop_mul,cSGMCMC}. Prior research has underlined the importance of capturing multiple modes in the posterior distribution for accurately estimating epistemic uncertainty \citep{cSGMCMC}. Therefore, to effectively address distribution shifts in the medical domain, a comprehensive examination of uncertainty estimation methods is essential, enabling the identification of the most trustworthy method for robust uncertainty assessment.

In this paper, we explore the utility of different methods to estimate uncertainty in distribution shift of medical images in segmentation task. We carefully select methods from diverse families, considering whether they capture one or multiple modes in the posterior distribution. Subsequently, we design various distribution shifts, both synthetic (e.g., injecting Gaussian noise, rotation, and corruption) and natural (e.g., different modalities), commonly encountered in the medical domain. Next, we conduct a comprehensive comparison of the aforementioned methods, employing different metrics to evaluate which model yields the most trustworthy and accurate uncertainty estimates across various shifts. Our findings strongly suggest that models capable of capturing \textit{multiple modes} in the posterior distribution provide more reliable uncertainty estimation. These results hold significant implications for improving uncertainty assessment in medical image segmentation tasks and enhancing the overall performance and trustworthiness of predictive models in real-world medical applications.

%Then we compare the aforementioned methods in terms of different metrics to explore which model produces the most trustworthy and accurate estimate of uncertainty across different shifts. Our findings indicate that models in which capture different modes in the posterior provide more reliable uncertainty estimation.
 
%Bayesian approaches are divided into tow big families: Variational Inference (VI) and Markov Chain Monte Carlo (MCMC) methods. Non-Bayesian approaches include training multiple probabilistic neural networks with ensembeling. While VI methods usually approximate a single mode in the posterior, MCMC methods are used for sampling from different modes. 

\paragraph{Contributions} In the quest to understand the importance of capturing multimodal posterior for accurately characterizing epistemic uncertainty, this paper conducts a thorough assessment of predictive uncertainty within the context of distributional shift. We evaluate the performance of both uni-modal and multimodal approaches across a diverse range of common distribution shifts in medical image segmentation tasks.    
Our study aims to address the following key questions: 
%\begin{enumerate}
%\item 
1) Do the models that characterize multiple modes in the posterior distribution produce more reliable predictive uncertainty? 
%\item 
2) How dose calibration in in-distribution (ID) setting correlate with calibration under dataset shift?
%\item 
%3) How do shifts in datasets impact the concurrent behavior of model uncertainty and performance?
3) How does the covariance between model uncertainty and performance alter under different distribution shifts?

%\end{enumerate}

\section{Related Works}\label{sec:related_work}
%First paragraph:papers that worked on dataset shift in medical domain 
%2d paragraph: About UQ for OOD detection: this paragraph should not only include BNN papaers but also other OOD detection methods for classification
%In 2d paragraph I should also mention the reason why I choose 3 methods
\paragraph{Uncertainty Estimation Methods}
A variety of methods have been developed for quantifying predictive uncertainty in DNNs. BNNs \citep{27,28} learn a posterior distribution over parameters that quantifies parameter uncertainty, a type of epistemic uncertainty that can be reduced through the collection of additional data. Sampling the posterior distribution poses challenges in general cases, leading to the adoption of approximation methods. Among these methods, Markov Chain Monte Carlo methods (MCMC) \citep{mcmc} stand out as a popular choice for accurately sampling the posterior distribution, while Variational Inference (VI) \citep{VI} offers a technique for learning an approximate posterior distribution. Non-Bayesian methods to uncertainty quantification include training multiple probabilistic neural networks with bootstrap or ensembling of networks, a technique commonly referred to as Deep Ensemble \citep{DE}. Popular approximate Bayesian approaches that fall in VI family include: Laplace approximation \citep{laplaceApp}, Stochastic Variational Bayesian Inference (SVI) \citep{VI,pVI}, dropout-based variational inference known as Monte Carlo Dropout (MCD) \citep{MCD} and SWAG \citep{SWAG}. These approaches are well-known that capture only \textit{single mode} in the posterior distribution, consequently underestimating uncertainties \citep{bishop}. From MCMC family, Stochastic Gradient MCMC methods (SG-MCMC) \citep{SGLD} was proposed which combine MCMC methods with minibatching, enabling scalable inference on large datasets. Cyclical Stochastic Gradient MCMC (cSG-MCMC) \citep{cSGMCMC}, combine SG-MCMC with cyclic learning rate schedule enabling to capture \textit{multimodal} posterior distribution within realistic computational budget. In another work, multiSWAG \citep{GPB}, combines SWAG with Deep Ensemble to capture different modes in the posterior. The authors provide a Bayesian perspective for Deep Ensemble in their paper, highlighting different training trajectories converge to different modes in the posterior distribution. However, our empirical results reveal that Deep Ensemble, in practice, tends to lack diversity among different models and fails to provide reliable uncertainty across different distribution shifts.
%consequently underestimates uncertainties.
%However, in our empirical results we observed that deep ensemble suffers from the lack of diversity among different models and underestimate uncertainties. 
%\paragraph{OOD Detection Methods}
%There is another line of research that they try to detect out-of-distribution samples, considering the output probability rather than inferring a distribution over parameters \citep{Gimpel, ODIN,mahalonobis}. These methods which are well-known as threshold-based detectors are not of our focus in this study.
% ,enscor,grammatrix
\paragraph{Using Distributional Shift to Evaluate Predictive Uncertainty}
Previous studies assess predictive uncertainty from various methods under dataset shift using standard benchmark datasets such as CIFAR10 and FashionMNIST \citep{trust, OOD_bench}.
%In another recent study, \citet{bench}, evaluates various Bayesian methods from VI family (plus deep ensemble) under two types of distribution shift in Diabetic Retinopathy classification task.
In a recent study, \citet{bench} evaluated a range of Bayesian methods from VI family and Deep Ensemble, in the context of two types of distribution shifts within the Diabetic Retinopathy classification task. In \citet{multihead}, multi-head Convolutional Neural Networks (CNNs) are proposed for predictive uncertainty estimation. The study compares its performance in Out-of-Distribution (OOD) detection with Deep Ensemble and MCD using a breast cancer dataset.
%In \citet{multihead}, authors propose multi-head CNNs to estimate predictive uncertainty and compare its performance on Out-of-Distrribution (OOD) detection with Deep Ensemble and MCD in breast cancer dataset.
%\citet{relOOD} investigate reliable machine learning methods in health domain under various distribution shifts using different input data types. However, they use some OOD detection methods that do not take into account model's predictive uncertainty.
\citet{relOOD} explore the reliability of machine learning methods in the health domain under various distribution shifts, utilizing different input data types. However, they employ OOD detection methods that do not consider the model's predictive uncertainty.
Finally, in \citet{Med_pred}, the authors assess uncertainty values derived from patch-based classifiers. They compare MCD, Deep Ensemble, and multi-head CNNs on digital pathology datasets to discriminate between in-distribution and OOD data.

%Our work differs from existing literature in multiple aspects.
Our work stands out from the existing literature in several key aspects. We are going to investigate the significance of effectively capturing diverse modes in the posterior for accurate epistemic uncertainty estimation in real-world settings. To achieve this objective, we focus on the distributional shift detection task. We have thoughtfully selected three distinct methods--cSGHMC, MCD, and Deep Ensemble-- spanning various families: MCMC, VI, and Deep Ensemble which are well-established, widely-used, and consistently outperform other techniques within their respective categories. Each method showcases the ability of capturing either single mode or different modes in the posterior.
%well-established and widely-used uncertainty quantification methods within the field of medical image analysis including cSGHMC, MCD and Deep Ensemble: from various families: MCMC, VI, and Deep Ensemble. 
In pursuit of our goal, we introduce a range of distribution shifts commonly encountered in medical domain and consider several challenging 3D medical datasets in segmentation task.  

\section{Background: Uncertainty Methods}
%In this section, we provide background on the uncertainty mehods that we aim to utilize in our study.
In this section, we provide background on the uncertainty estimation methods we intend to employ in our study. 
%We aim to leverage state-of-the-art Bayesian and non Bayesian methods in out-of-distribution (OOD) detection task in the health domain. This will enable us to determine which method provides more accurate uncertainty estimation in OOD detection.
%We have carefully chosen three distinct methods from various families: MCMC, VI, and Deep Ensemble. These methods are well-established, widely-used, and consistently outperform other techniques within their respective categories.% Our goal is to highlight the strengths and weaknesses of each method from various families, enabling practitioners to readily adopt the most appropriate approach for safely utilizing deep learning models in healthcare.      

 %\subsection{Maximum A Posteriori Estimation (MAP)}

 \subsection{Bayesian Neural Networks}
 BNNs estimate the posterior distribution over unknown parameters rather than estimating a single value. Let assume $\mathcal{D}:=\{(x_{i}, y_{i}), i=1,\dots, n\}$ be a dataset containing $n$, i.i.d. samples. Given a prior distribution $p(\theta)$ on the network parameters, the Bayes rule allows to estimate the posterior distribution as: $p(\theta| \mathcal{D}) \propto p(\mathcal{D} | \theta) p(\theta)$. Given $p(\theta | \mathcal{D})$, a predictive distribution for a new instance $x^{*}$ is obtained by marginalizing over all possible model parameters:  
 \begin{equation}
\label{marginal}
p(y^{*}|x^{*},\mathcal{D}) = \int p(y^{*}|x^{*},\theta) p(\theta|\mathcal{D}) d\theta 
\approx \frac{1}{S} \sum_{s=1}^{S} p(y^{*}|x^{*},\theta_{s}), \theta_{s} \sim p(\theta|\mathcal{D}). \nonumber
\end{equation}
The model’s predictive uncertainty for a new instance $x^{*}$ can be obtained using the predictive entropy:
\begin{equation}
\mathcal{H}(y^{*}|x^{*}, \mathcal{D})= -\sum_{c\in C} p(y^{*}|x^{*}, \mathcal{D}) \log p(y^{*}|x^{*}, \mathcal{D}),
\end{equation}
where $C$ is the number of distinct classes. 

Computing a posterior distribution over the parameters of a neural network using Bayes' rule is analytically intractable, necessitating the adoption of approximate inference methods \citep{cSGMCMC,MCD,DE}. Below, we describe state-of-the-art methods with unimodal or multimodal posterior distribution.
  
 \subsection{Cyclical Stochastic Gradient Hamiltonian Monte Carlo (cSGHMC)}
The idea behind MCMC methods is to construct a Markov
chain of samples which eventually converges to a desired distribution \citep{bench}. SGHMC couples a SGD optimizer with Langevin Dynamics to make a Markov chain algorithm to sample from the posterior $p(\theta|\mathcal{D})$ \citep{SGLD}. Posterior samples are updated at the $k$-th step as: $\theta_{k} = \theta_{k-1}+m_{k-1}$, $m_{k} = \beta m_{k-1} -\frac{\ell_{k}}{2}n\nabla\tilde{U}(\theta_{k})+ \sqrt{(1-\beta)\ell_{k}}\epsilon_{k}$, where $\beta$ is the momentum term, $\ell_{k}$ is the stepsize, and $\epsilon_{k}$ has a standard Gaussian distribution. $\tilde{U}(\theta)=\frac{1}{N} \sum_{i=1}^{N} \log p(x_{i}|\theta)+\frac{1}{n}\log p(\theta)$, is an unbiased estimator of the log posterior $U(\theta)= \log p(\mathcal{D}|\theta) + \log p(\theta) $ over a mini-batch of size $N$. 
%This allows applying MCMC methods on large scale datasets.
%The method enables applying MCMC methods on large scale datasets when dataset has to split into mini-batches. 
To guarantee the convergence to the true posterior, the stepsize $\ell_{k}$ should 
%satisfy Robbins-Munro conditions \citep{SGLD}.
be reduced towards zero while ensuring that $\sum_{k=0}^{\infty} \ell_{k} = \infty$ \citep{SGLD}. 
%The  true Bayesian posterior is a complex multimodal distribution. 
cSGHMC enhances posterior exploration by using a cyclical learning rate schedule with warm restart, where the algorithm is restarted with a large learning rate $\ell_{0}$ and subsequently reduced to a small value within a cycle of $L$ iterations.
%restarting the algorithm with a large learning rate $\ell_{0}$ and lowering it to a small value within a cycle of $L$ iteration.
Using warm restart helps to mitigate the mode collapse problem often encountered in VI methods and allows for capturing different modes in the posterior. This enables a better estimate of the epistemic uncertainty when processing unseen data \citep{cSGMCMC, handson}. 
 
 \subsection{Monte Carlo Dropout (MCD)}
 Variational inference is an approximate inference method that seeks to approximate the posterior distribution with a simpler distribution $q_{\phi}(.)$ that is parametrized by a set of parameters $\phi$.
%Variational inference is an approximate inference method that seeks to sidestep the intractability of exact posterior inference over the network parameters by framing posterior inference as a variational optimization problem \citep{bench}.
In particular, variational inference in neural networks seeks to find an
approximation to the posterior distribution by solving the optimization problem: $\text{argmax}_{q\in Q} \{-D_{\text{KL}} (q||p)+\ln p(y|x,\theta) \}$ where $Q$ is a variational family of distributions and $p$ is a prior distribution \citep{bench}.

\citet{MCD} demonstrated that training a deterministic neural network with $\ell_2$ and dropout regularization, yields as approximation to variational inference in a BNN \citep{bench}. In this method, to sample from the approximate posterior distribution, dropout is turned on at evaluation time, which results in a distribution for the output predictions \citep{handson}. However, it is important to note that despite its benefits, variational inference can suffer from sample overlay, even within a mode, potentially limiting its ability to accurately capture uncertainty.

%The main idea behind VI is to approximate the posterior distribution by minimizing the Kullback-Leibler divergence (Kullback and Leibler, 1951) between the posterior and a simpler distribution qϕ(.), that is parametrized by a set of parameters ϕ (Jospin et al., 2020).

 \subsection{Deep Ensemble (DE)}
A deep ensemble as introduced by \citep{DE} comprises multiple deterministic networks, each independently trained from scratch with different random seeds. In contrast to BNNs, deep ensemble does not infer a distribution over the parameter of a neural network. Instead, it establishes a predictive distribution by marginalizing over multiple deterministic networks.
%Instead, it achieves predictive distribution by marginalizing over multiple deterministic networks. 
While traditionally categorized as non-Bayesian methods, \citet{GPB} shed light on the Bayesian characteristic of deep ensembles. They argue that training a deterministic model with regularization corresponds to identifying the modes of a Bayesian posterior.% a phenomena that is against our empirical results.

%they marginalize over multiple deterministic networks to obtain a predictive distribution and estimating uncertainty. They are known as non-Bayesian methods, however \citet{GPB} demonstrate a Bayesian characteristic of this method, by arguing that training a deterministic model with regularization corresponds to locating the modes of a Bayesian posterior.  

%While commonly regarded as non-Bayesian, this method was demonstrated to exhibit characteristics of Bayesian behavior, as training a deterministic model with regularization corresponds to locating the modes of a Bayesian posterior \citep{GPB}.

\subsection{Maximum A Posteriori Estimation (MAP)}
As an alternative to inferring a posterior distribution over neural network parameters, maximum a posteriori (MAP) estimation provides a point estimate of the network parameters which is equal to the mode of the exact posterior distribution. Notably, if the regularizer imposed on the model parameters is interpreted as the log of a prior on parameters distribution, optimizing the cost function can be seen as a MAP estimate of model parameters. We term this approach as MAP and we consider, its softmax entropy as a measure of uncertainty.

%MAP estimation, an alternative to inferring a posterior distribution over neural network parameters, yields values equal to the mode of the exact posterior distribution. We refer to this method as MAP and use softmax entropy as a measure of uncertainty.

\section{Experimental Setup}
\label{sec:data}
In this section, we present the datasets and distribution shifts considered in our study.
\subsection{Datasets}
We utilize three different 3D medical datasets: Hippocampus \citep{Hipoc}, AMOS2022 \citep{Amos} and KiTS21 \citep{Kits}. %In this section, we will provide a thorough explanation of the datasets and a detailed description of our process for creating the shifted distribution test samples.
%We have selected distribution shifts that consist of both near- and far-from-distribution samples that represent realistic use cases in the real-world settings.
%We consider distributionally shifted test set caused by covariate shifts, modality shift, and corruption.
For our study, we introduce distributionally shifted test sets, induced by covariate shifts, modality shifts, and corruption.
The covariate shifts and data corruption aim to assess a model's predictive uncertainty when tested on data related to the same classification task but originating from different data sources and environmental conditions. By modality shift we aim to assess the models under natural distribution shift.  

\paragraph{Hippocampus}
dataset was collected from Vanderbilt University Medical Center, USA. The primary objective is to segment the left and right parts of the Hippocampus. Each scan in this dataset is annotated with 3 different classes: background (class 0), anterior (class 1) and posterior (class 2).
\begin{itemize}
\item 
\textbf{In Distribution}
The dataset consists of 260, 3D labeled MRI scans. We split the dataset into three parts: the training set, the validation set, and the test set, each containing 210, 25, and 25 images, respectively. In order to create the distributionally shifted test set we apply two types of synthetic shifts on test set.
\item 
\textbf{Distribution shift}
Adding \textit{Gaussian noise} with different intensities ranging from: 2, 4, 6, 8, 10, 12, to 14 to the clean test set. It creates seven distributionally shifted test sets each containing 25 scans. 
\item 
\textbf{Distribution shift}
\textit{Rotating} the images with different angles ranging from: 10, 20, 30, 40, 50, 60, 70, 80, to 90. This generates a total of seven distributionally shifted test sets each containing 25 scans. 
An illustration of ID and distributionally shifted images along with their kernel density estimate (KDE) plots for both distribution shifts are provided in Appendix A, Figures \ref{Fig:cor_hippo} and \ref{Fig:kde_hippo}.
\end{itemize}

\paragraph{AMOS}
is a large-scale dataset for abdominal organ segmentation with voxel-level annotations for 15 organs. %The organs include the spleen, right kidney, left kidney, gallbladder, esophagus, liver, stomach, aorta, inferior vena cava, pancreas, right adrenal gland, left adrenal gland, duodenum, bladder, and prostate/uterus.
The dataset consists of 600 subjects, 
%at the Longgang District People’s Hospital in China
including 500 CT and 100 MRI scans. We frame the problem as the multi-class classification with 15 class labels. 
\begin{itemize}
\item 
\textbf{In Distribution}
We use 200 CT scans as our ID dataset to train the model. The dataset is further split into training, validation and test sets each containing 180, 10 and 10 scans respectively.  
\item 
\textbf{Distribution shift}
We use MRI scans as distributionally shifted dataset. This provides a test set with different modality, resembling a natural distribution shift. An illustration of the ID and distributionally shifted scans along with their KDE plots are provided in Appendix B, Figure \ref{Fig:amos_kde}. 
\end{itemize}
\paragraph{KITS} 
is a dataset designed for semantic segmentation of the kidney tumors and their host kidneys. The data set includes 300 3D images along with their segmentation mask of tumors, kidneys, and cysts. We combine tumors and cysts in one class; therefore, each scan in this dataset has 3 different classes: background (class 0), kidney (class 1) and tumor+cyst (class 2).
%The dataset was collected from M Health Fairview, Minnesota, United States, and Cleveland Clinic, United States.

\begin{itemize}
\item 
\textbf{In Distribution}
We use the original dataset as ID and split it into three parts: training, validation and test sets, containing 200, 40, and 59 scans, respectively.
\item 
\textbf{Distribution shift}
To generate a distributionally shifted test set, we introduce corruption to the test images by incorporating rectangles filled with zeroes at random positions within each scan. %Further details regarding the corruption process, along with 
Illustrations of ID images, corrupted images, and their corresponding KDE plots are provided in Appendix C, Figure \ref{Fig:kits_kde}. 
\end{itemize}

\subsection{Evaluation Metrics}
For our evaluation study, we employ two types of metrics: calibration metrics and metrics designed to assess the predictive uncertainty.
\paragraph{Calibration}
Commonly used metrics to evaluate the quality of model uncertainty include: Negative Log Likelihood (NLL) \citep{NLL}, Brier Score (BS) \citep{BS} and Expected Calibration Error (ECE) \citep{ECE}. NLL is a popular metric for evaluating predictive uncertainty \citep{DE}.
%measuring the quality of in-domain uncertainty of deep learning models.
It directly penalizes high probability scores assigned to incorrect labels and low probability scores assigned to the correct labels. BS is computed as the squared error of a predicted probability vector and the one-hot encoded true label. ECE measures the correspondence between predicted probabilities and empirical accuracy.
\paragraph{Uncertainty Evaluation}
In order to investigate how a model’s predictive uncertainty estimates vary across different methods, we plot histogram of predictive uncertainty (PDF) for distributionally shifted samples. A model that produces reliable uncertainty estimates should assign high uncertainty to samples that it classifies incorrectly or samples with distribution different from training data. Hence, the mass of distribution should concentrate at a higher value. %Additionally, we plot an empirical cumulative distribution function (CDF) for the entropy of the predictive distributions on shifted samples. A method that demonstrates a lower probability for a low entropy prediction is considered more effective in capturing uncertainty.

\subsection{Implementation details}
In this section, we provide thorough details on the implementation of all methods and uncertainty estimation. For training all models we adopt U-Net architecture \citep{unet} with parameters reported in Appendix D and the Cross Entropy as the loss function\footnote{For a well-calibrated uncertainty estimation, it is essential to use a proper scoring rule \citep{DE}.}. We utilize a mini-batch of size $2$ and apply feature scaling and resizing for both train and test sets. To train cSGHMC, we utilize a SGHMC optimizer with a cyclic learning rate schedule \citep{cSGMCMC}. For other methods, we employ a SGD optimizer with momentum of $0.99$ and a fixed learning rate. Additional implementation details can be found in Appendix D. %\ref{app:D}. 
To generate a predictive distribution using cSGHMC, we draw $20$ samples from the final $20$ cycles during training, selecting each sample at the end of the cycle where the loss is minimum. For MCD, we obtain $20$ samples through forward pass during evaluation whit dropout activated. Following \citet{Ass_Unc_ID}, we implement dropout after each convolution layer. For determining dropout rate  a careful consideration is necessary; as an increased dropout rate enhances stochasticity but may adversely affect performance (see Table \ref{table:OOD} in Appendix D). In our setting, we adjusted the dropout rate to a level that dose not significantly diminish performance. Specifically, for the Hippocampus dataset, we increased the dropout rate to (p=$0.2$), and for the AMOS and KITS datasets, we set it to (p=$0.05$). As for DE, we train $20$ deterministic models from scratch, each with a different random seed. All training is done in a server (Intel Xeon 2.1GHz, 64GB, GeForce RTX 2080 Ti) from an internal cluster.

To assess uncertainties, we utilize 20 network samples of each method. We calculate the class probability $p_{c}= \frac{1}{20}\Sigma_{s=1}^{20}p_{c,s}$, where $p_{c,s}$ represents the softmax output for class $c$ in network sample $s$, and $C$ is the set of classes.  Subsequently, we compute the entropy $\mathcal{H}=-\Sigma_{c\in C} p_{c}\log p_{c}$ based on these probabilities, serving as a measure of uncertainty. The computation of voxel-wise uncertainties results in a 3D uncertainty map for each image. Aggregating uncertainties for each image involves calculating the average over all voxels in the prediction mask classified as True Positive (TP), False Positive (FP), or False Negative (FN). Notably, True Negative (TN) voxels are excluded from our evaluation, as they contribute minimal uncertainty and have the least impact on our study. This exclusion aims to minimize the background effect, which is significant in medical images. %We present the results in the next section. 
%We extend our analysis by computing uncertainties on both in distribution and distributionally shifted test sets, and the results are presented in the next section.
%We then ran inference on in and out-of-distribution datasets for the models. We report performance in Table \ref{table:per_imp}. 
%\newpage
\section{Main Results}
In this section, we present the main findings related to performance, calibration, and uncertainty quantification across diverse methods and distribution shifts. Additionally, we provide insightful qualitative results, including uncertainty maps for each method, spanning diverse distribution shifts in Appendix G. 
%First, we will present the results on Hippocampus dataset.  

%\begin{table}[!t]
%\caption{Performance impact by distribution shift}  %Results obtained from our method are underlined and the best results are indicated in bold.} 
%\centering
%\begin{tabular}{|l| c | c c c c|}
%\hline
%\textbf{Data} & \textbf{In-Distribution} & \multicolumn{4}{c|}{\textbf{Distribution Shift}} \\
%\hline
%Hippocampus &  & \multicolumn{2}{c}{\textit{Gaussian noise}} & \multicolumn{2}{c|}{\textit{Rotation}} \\
% & $88\%$ & int:2 & int:6 & ang:10 & ang:30 \\
% &  &  $81\%$ & $44\%$ & $80\%$ & $58\%$\\
%\hline
%AMOS &  & \multicolumn{4}{c|}{\textit{Different modalities}} \\
%   & $87\%$ & \multicolumn{4}{c|}{$23\%$} \\
%\hline
%KIT &  & \multicolumn{4}{c|}{\textit{Corruption}}\\
%    &   $93\%$   & \multicolumn{4}{c|}{$30\%$}\\
%\hline
%\end{tabular}
%\label{table:per_imp}
%\end{table}

\subsection{Covariate Shift}
As stated in prior works, we expected the performance of a model to degrade as it predicts on increasingly shifted data and ideally this reduction in performance would coincide with increased entropy \citep{trust}. In terms of calibration, we expect a model that is well-calibrated on training data remains calibrated on shifted data too \citep{trust}. 

\begin{figure}
\setkeys{Gin}{width=1\linewidth}
%\begin{minipage}[t]{0.24\textwidth}
\begin{subfigure}{0.236\textwidth}
    \includegraphics{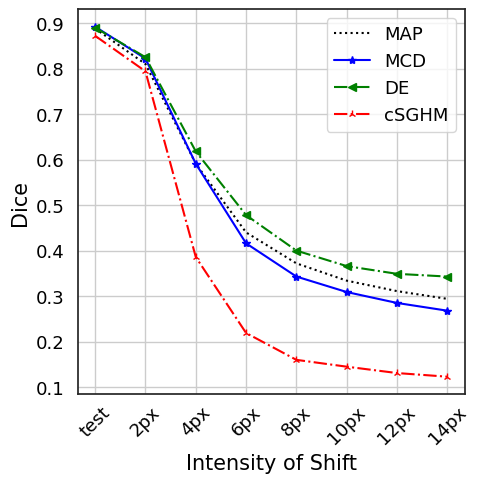}
%\subcaption{Dice score} \label{fig:dice}
%\end{minipage}\hfill
     \caption{Dice score} \label{fig:dice}
\end{subfigure}\hskip1ex
%\begin{minipage}[t]{0.24\textwidth}
\begin{subfigure}{0.24\textwidth}
    \includegraphics{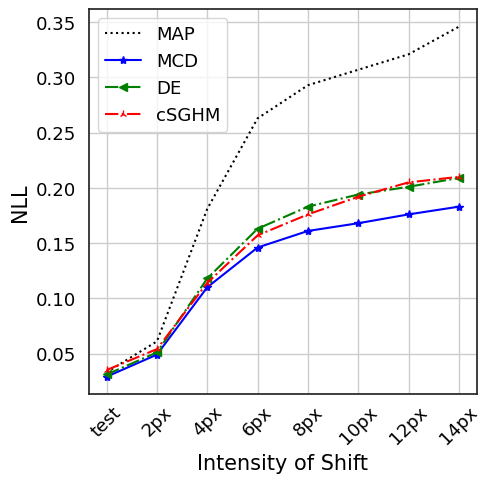}
    \caption{NLL} \label{fig:nll}
%\end{minipage}\hfill
\end{subfigure}\hskip1ex
%\begin{minipage}[t]{0.24\textwidth}
\begin{subfigure}{0.24\textwidth}
  \includegraphics{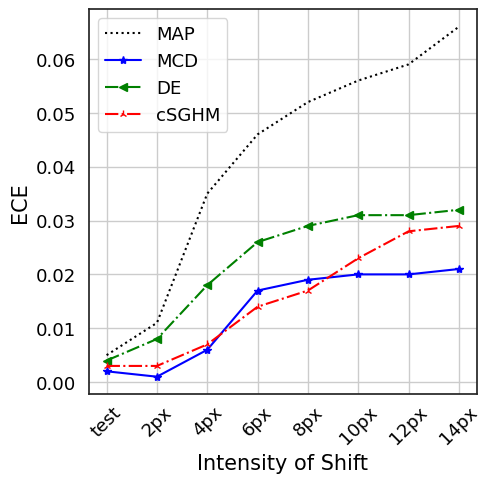}
  \caption{ECE} \label{fig:ece}
%\end{minipage}\hfill
\end{subfigure}\hskip1ex
%\begin{minipage}[t]{0.24\textwidth}
\begin{subfigure}{0.24\textwidth}
   \includegraphics{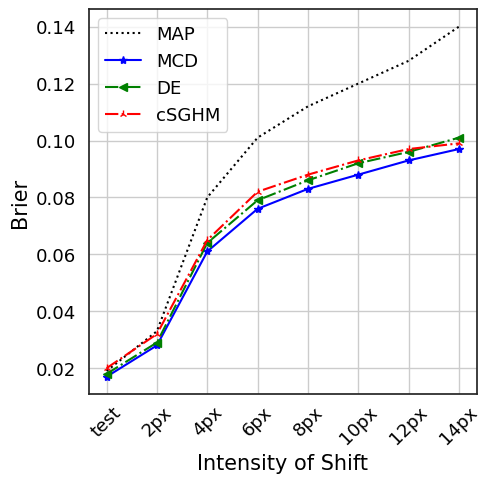}
%\subcaption{Brier} \label{fig:brier}
\caption{BS} \label{fig:brier}
\end{subfigure}
\\
%\end{figure}
%\begin{figure}
%\setkeys{Gin}{width=1\linewidth}
%\label{Fig:unc_hipo}
%\begin{minipage}[t]{0.24\textwidth}
\begin{subfigure}{0.24\textwidth}
   \includegraphics{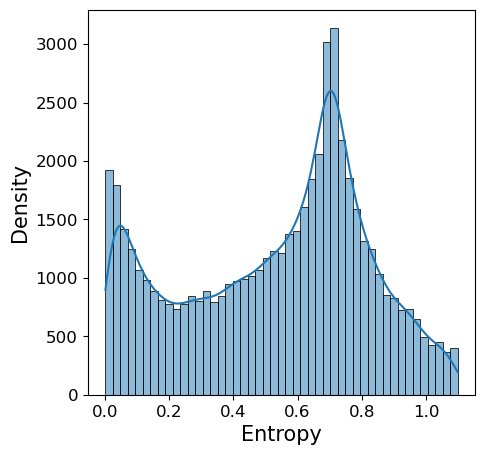}
   \caption{cSGHMC} \label{fig:mcmchip}
\end{subfigure}\hskip1ex
%\end{minipage}\hfill
%\begin{minipage}[t]{0.24\textwidth}
\begin{subfigure}{0.24\textwidth}
   \includegraphics{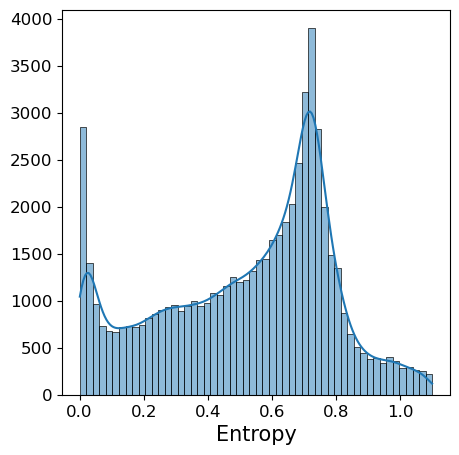}
   \caption{MCD} \label{fig:mcdhip}
%\end{minipage}\hfill
\end{subfigure}\hskip1ex
%\begin{minipage}[t]{0.24\textwidth}
\begin{subfigure}{0.24\textwidth}
   \includegraphics{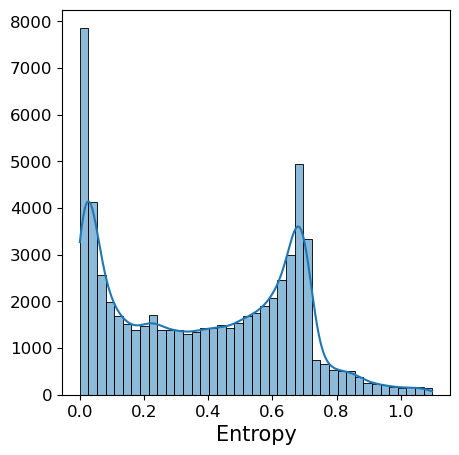}
    \caption{DE} \label{fig:dehip}
%\end{minipage}\hfill
\end{subfigure}\hskip1ex
%\begin{minipage}[t]{0.24\textwidth}
\begin{subfigure}{0.24\textwidth}
   \includegraphics{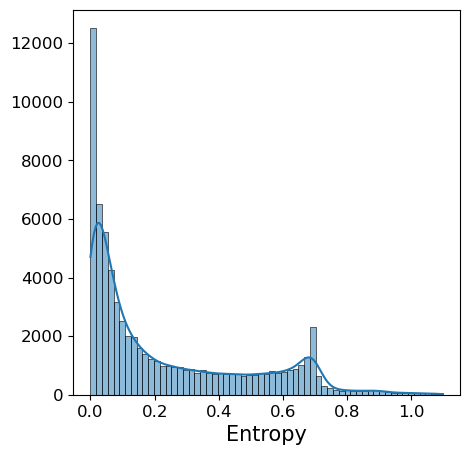}
   \subcaption{MAP} \label{fig:maphip}
%\end{minipage}
\end{subfigure}\hskip1ex
\caption{Performance and Calibration Metrics Comparison on Blurred Images. The first row indicates performance and various calibration metrics. The second row presents the histogram of predictive uncertainty across different models for shift equal to 6px, with cSGHMC and MCD exhibiting the highest uncertainty.}
\label{Fig:Hipo}
\end{figure}

The results on Hippocampus dataset are visualized in Figure \ref{Fig:Hipo}. As depicted in Figure \ref{fig:dice}, the Dice score degrades with increasing shift for all methods, with the cSGHMC method exhibiting a more pronounced decline in prediction quality. %While cSGHMC achieves the worst accuracy on the test set, it actually outperforms all other methods by a much larger margin when exposed to significant shift.
In Figure \ref{Fig:Hipo} (b-d) we look at the calibration of different methods. One observation is that while all the methods are well calibrated in terms of all metrics on test set, miscalibration increases as the intensity of shift increases. This means that calibration on ID dose not necessarily lead to calibration on shifted data. Second observation is that all marginalising methods outperform MAP estimation by large margin. The third observation is that while all marginalizing methods perform almost the same in terms of BS, MCD outperforms the other two methods in terms of NLL and ECE under severe shift (\textit{shift} > 8px). In Figure \ref{Fig:Hipo} (e-h), wee look at the histogram of the predictive entropy of different methods where intensity of shift is equal to $6$ pixel. We observe that cSGHMC and MCD assign high entropy to shifted data, while DE and MAP give low entropy and high confidence prediction on shifted data, i.e. they are confidently wrong about shifted data. 

\begin{figure}[!h]
\setkeys{Gin}{width=1\linewidth}
%\begin{minipage}[t]{0.24\textwidth}
%\begin{subfigure}[t]{0.24\textwidth}
\begin{subfigure}{0.24\textwidth}
    \includegraphics{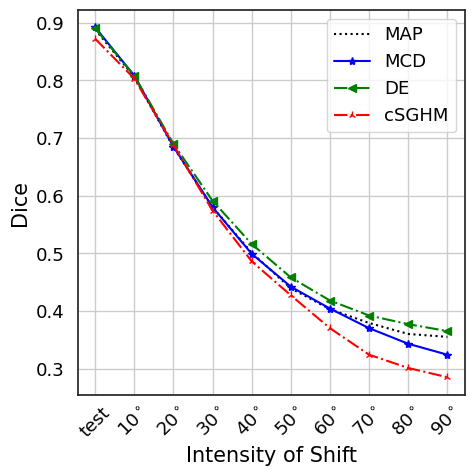}
    \caption{Dice score} \label{fig:rot_dice}
\end{subfigure}\hskip1ex
%\begin{minipage}[t]{0.24\textwidth}
\begin{subfigure}{0.24\textwidth}
    \includegraphics{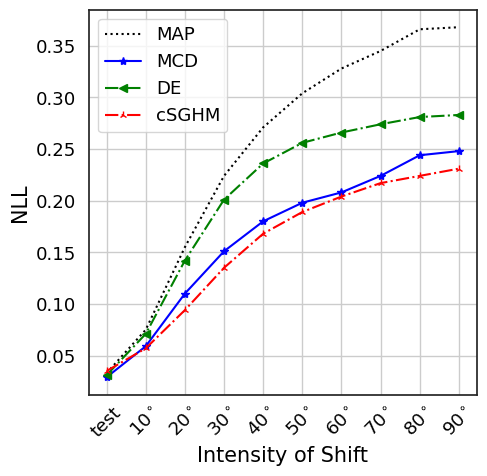}
    \caption{NLL} \label{fig:nll_rot}
%\end{minipage}\hfill
\end{subfigure}\hskip1ex
%\begin{minipage}[t]{0.24\textwidth}
\begin{subfigure}{0.24\textwidth}
   \includegraphics{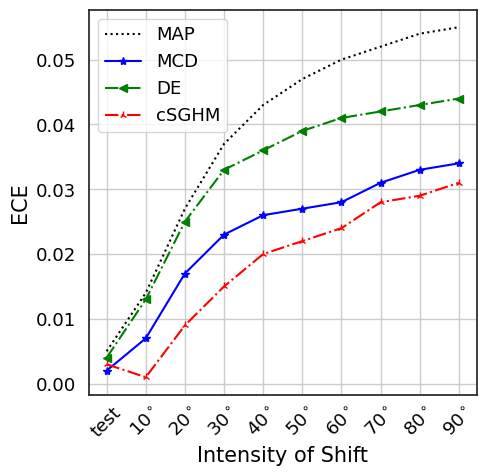}
   \caption{ECE} \label{fig:rot_ece}
%\end{minipage}\hfill
\end{subfigure}\hskip1ex
%\begin{minipage}[t]{0.24\textwidth}
\begin{subfigure}{0.24\textwidth}
    \includegraphics{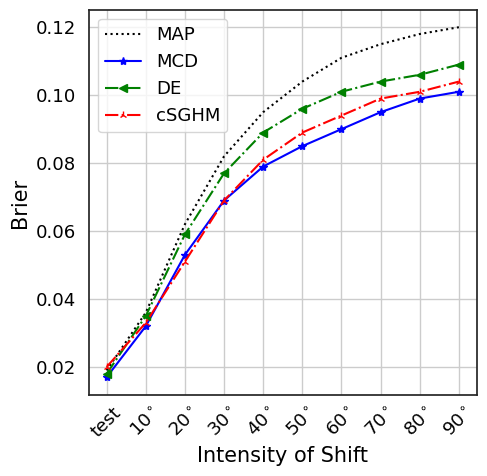}
    \caption{BS} \label{fig:rot_brier}
\end{subfigure}\hskip1ex
%\end{minipage}
%\end{figure}
\\
%\begin{figure}[!h]
%\setkeys{Gin}{width=1\linewidth}
%\label{Fig:unc_rot}
%\begin{minipage}[t]{0.24\textwidth}
\begin{subfigure}{0.24\textwidth}
    \includegraphics{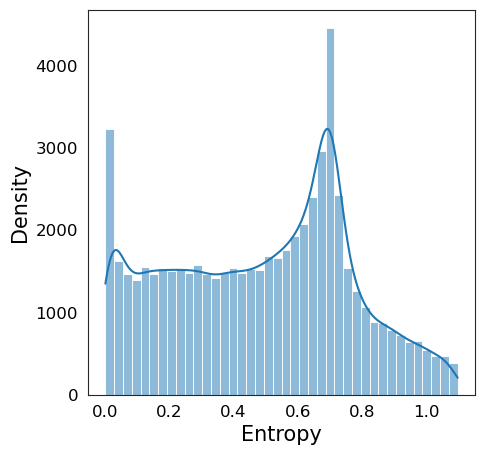}
    \caption{cSGHMC} \label{fig:rot_mcmc_unc}
%\end{minipage}\hfill
\end{subfigure}\hskip1ex
%\begin{minipage}[t]{0.24\textwidth}
\begin{subfigure}{0.24\textwidth}
   \includegraphics{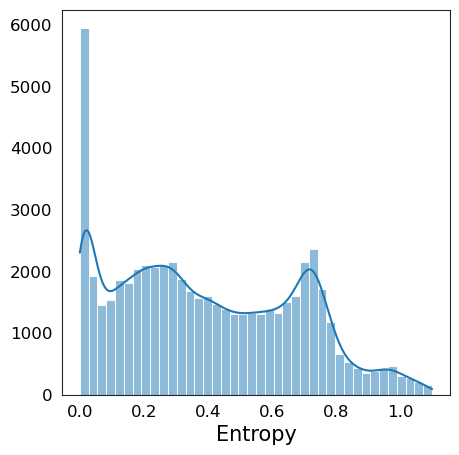}
   \caption{MCD} \label{fig:rot_mcd_unc}
\end{subfigure}\hskip1ex
%\end{minipage}\hfill
%\begin{minipage}[t]{0.24\textwidth}
\begin{subfigure}{0.24\textwidth}
    \includegraphics{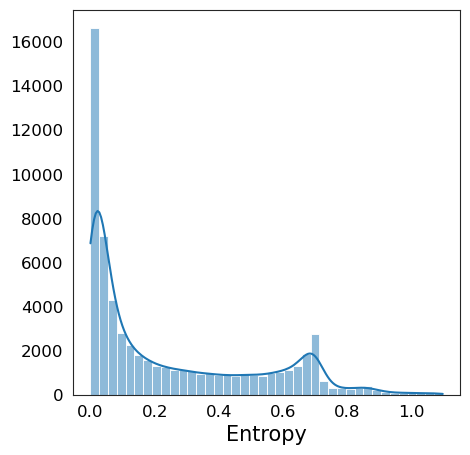}
    \caption{DE} \label{fig:rot_enc_unc}
%\end{minipage}\hfill
\end{subfigure}\hskip1ex
%\begin{minipage}[t]{0.24\textwidth}
\begin{subfigure}{0.24\textwidth}
    \includegraphics{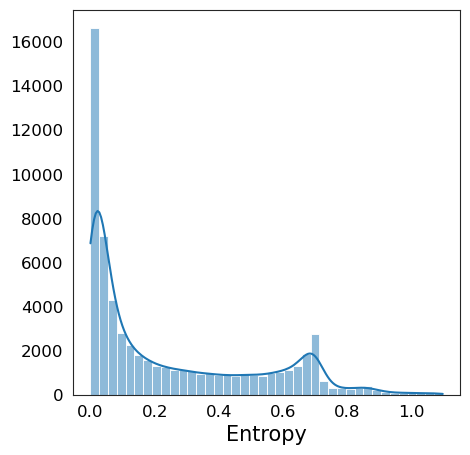}
    \caption{MAP} \label{fig:rot_map_unc}
%\end{minipage}
\end{subfigure}\hskip1ex
\caption{Performance and Calibration Metrics Comparison on Rotated Images. The first row presents a comparison of performance and various calibration metrics. The second row illustrates uncertainty estimation across different models for shift equal to $30^\circ$, with cSGHMC exhibiting the highest uncertainty}
\label{Fig:rot}
\end{figure}

In Figure \ref{Fig:rot} we present the results obtained under rotational shifts. Consistent with our earlier findings, all methods exhibit diminished performance and calibration as the degree of rotation increases. Notably, when comparing the performance across methods, cSGHMC shows a lower Dice score, particularly for significant shifts ( \textit{shift} > $50^{\circ}$). A closer look at calibration metrics reveals that DE performs worse than MCD and cSGHMC across all metrics. While MCD and cSGHMC demonstrate comparable BS, cSGHMC achieves lower NLL and ECE than other methods. Figure \ref{Fig:rot} (e-h), provides a visual representation of the predictive uncertainty histograms for rotated images with a $30^{\circ}$ angle. Notably, cSGHMC assigns the highest entropy to shifted data, while DE assigns very low entropy.  

\subsection{Modality Shift}
\begin{figure}[!h]
\setkeys{Gin}{width=1\linewidth}
\begin{subfigure}{0.24\textwidth}
    \includegraphics{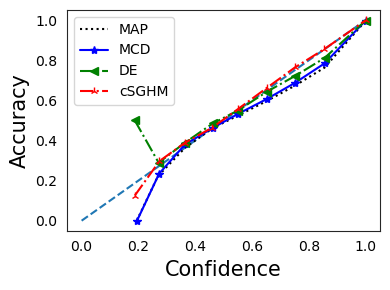}
    \caption{In Distribution} \label{fig:a}
\end{subfigure}\hskip1ex
\begin{subfigure}{0.24\textwidth}
    \includegraphics{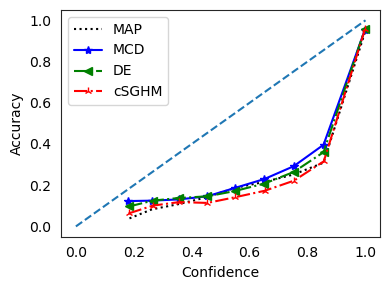}
    \caption{Distributional Shift} \label{fig:b}
\end{subfigure}\hskip1ex
\begin{subfigure}{0.24\textwidth}
    \includegraphics{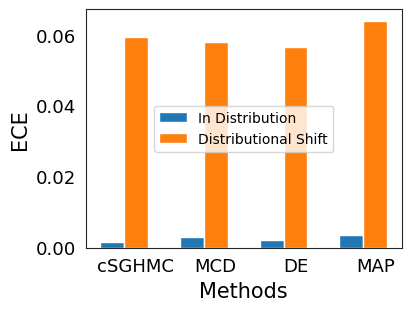}
    \caption{ECE} \label{fig:c}
\end{subfigure}\hskip1ex
\begin{subfigure}{0.24\textwidth}
    \includegraphics{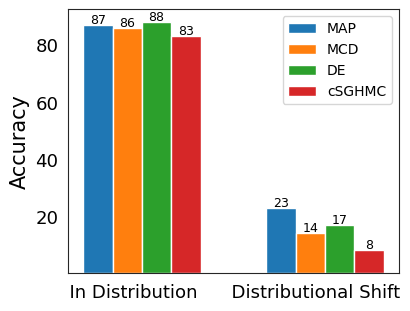}
    \caption{Performance} \label{fig:d}
\end{subfigure}\hskip1ex
%\end{figure}
\\
%\begin{figure}[!h]
%\setkeys{Gin}{width=1\linewidth}
\begin{subfigure}{0.24\textwidth}
    \includegraphics{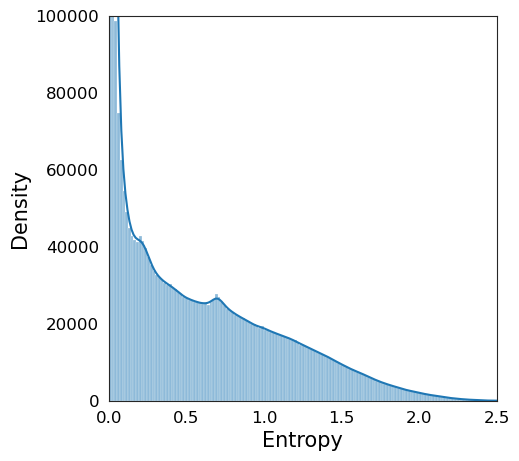}
    \subcaption{cSGHMC} \label{fig:a_unc}
\end{subfigure}\hskip1ex
\begin{subfigure}{0.24\textwidth}
    \includegraphics{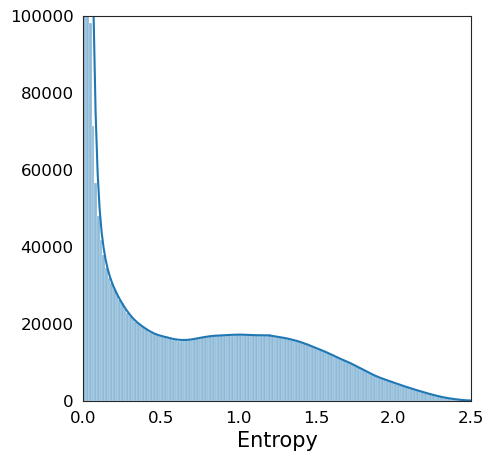}
    \caption{MCD} \label{fig:b_unc}
\end{subfigure}\hskip1ex
\begin{subfigure}{0.24\textwidth}
    \includegraphics{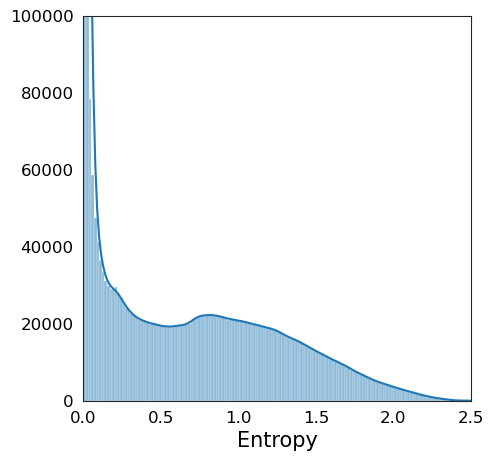}
    \caption{DE} \label{fig:c_unc}
\end{subfigure}\hskip1ex
\begin{subfigure}{0.24\textwidth}
    \includegraphics{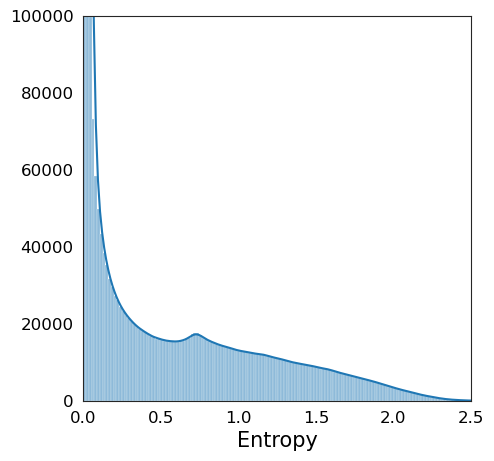}
    \caption{MAP} \label{fig:d_unc}
\end{subfigure}
\caption{Performance and Calibration Metrics Comparison on AMOS dataset. First row includes, reliability diagram, ECE and performance metrics. The second row depicts the histograms of
predictive uncertainty across various methods, with DE and cSGHMC exhibiting slightly higher uncertainty.}
%with all methods exhibiting nearly identical levels of uncertainty.
\label{Fig:amos}
\end{figure}

%We evaluate the performance of different methods under modality shift.
We assess the performance of various methods in the context of modality shift. In this task, models are trained on CT scans and evaluated on MRI scans, and the results are presented in Figure \ref{Fig:amos}. 
%In this task, models are trained on CT scans and evaluated on MRI scans. Results are depicted in Figure \ref{Fig:amos}. 
Despite the permanence drop under modality shift, none of the methods indicate high entropy on shifted data. Nevertheless, DE and cSGHMC assign higher entropy than the other two methods on shifted data. In terms of calibration, our observations indicate that all methods are well-calibrated when applied to ID data, with cSGHMC demonstrating superior calibration. However, when applied to shifted images, all methods exhibit significant miscalibration with DE showing a lower ECE. We refer to Appendix E for additional results; Table \ref{table:cal} reports further metrics such as NLL and BS. 

% performnace on KITS dataset for corruption 
\subsection{Corruption}
\begin{figure}[!h]
\setkeys{Gin}{width=1\linewidth}
\begin{subfigure}{0.24\textwidth}
    \includegraphics{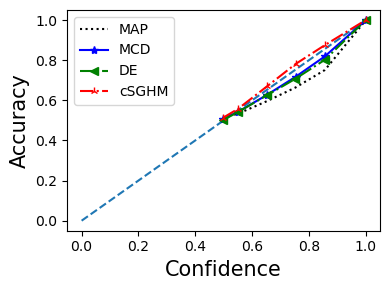}
    \caption{In Distribution} \label{fig:a_k}
\end{subfigure}\hfill
\begin{subfigure}{0.24\textwidth}
    \includegraphics{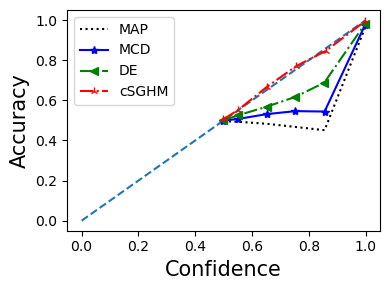}
    \caption{Distributional Shift} \label{fig:b_k}
\end{subfigure}\hfill
\begin{subfigure}{0.24\textwidth}
    \includegraphics{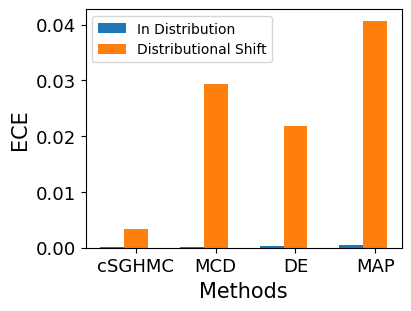}
    \subcaption{ECE} \label{fig:c_k}
\end{subfigure}\hfill
\begin{subfigure}{0.24\textwidth}
    \includegraphics{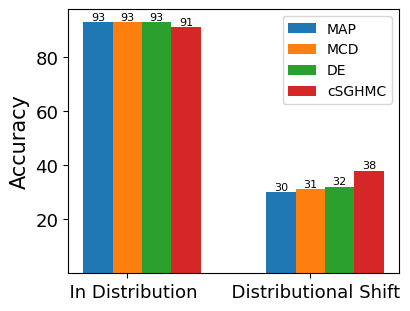}
    \caption{Performance} \label{fig:d_k}
\end{subfigure}\hfill
%\end{figure*}
\\
%\begin{figure}[!h]
%\setkeys{Gin}{width=1\linewidth}
\begin{subfigure}{0.24\textwidth}
    \includegraphics{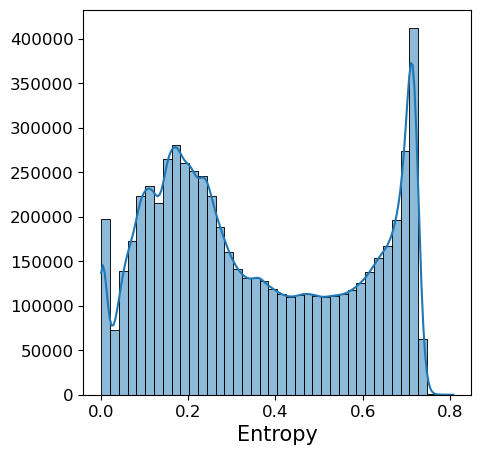}
    \caption{cSGHMC} \label{fig:a_ku}
\end{subfigure}\hfill
\begin{subfigure}{0.24\textwidth}
    \includegraphics{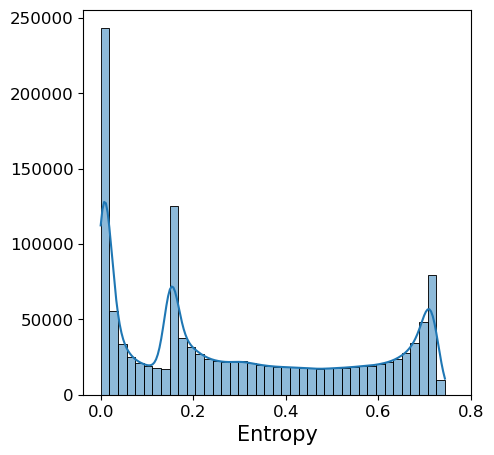}
    \caption{MCD} \label{fig:b_ku}
\end{subfigure}\hfill
\begin{subfigure}{0.24\textwidth}
    \includegraphics{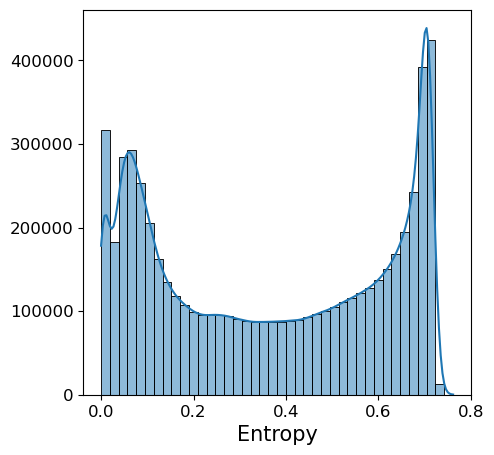}
    \subcaption{DE} \label{fig:c_ku}
\end{subfigure}\hfill
\begin{subfigure}{0.24\textwidth}
    \includegraphics{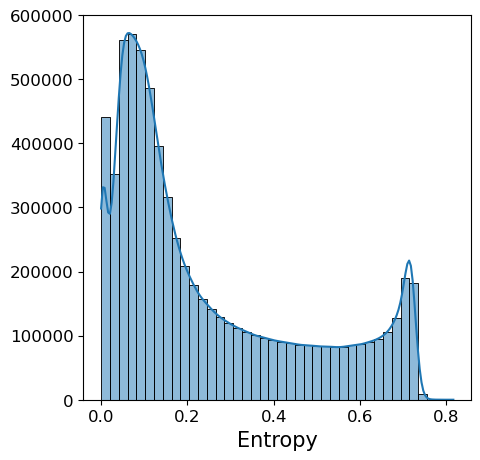}
    \subcaption{MAP} \label{fig:d_ku}
\end{subfigure}
\caption{Performance and Calibration Metrics Comparison on KITS dataset. First row encompasses, reliability diagram, ECE and performance metrics. The second row depicts the histograms of
predictive uncertainty across different methods, highlighting cSGHMC as the most calibrated with the highest uncertainty.}
\label{Fig:KIT}
\end{figure}

Finally, we assess the performance of various methods under corruption shift as detailed in Section \ref{sec:data}. For this task, the models are trained on the training set and subsequently evaluated on corrupted images. The results are visualized in Figure \ref{Fig:KIT}. Upon inspecting reliability diagrams, it becomes evident that all methods maintain good calibration for ID data. However, under distributional shift, all methods exhibit miscalibration except for cSGHMC (refer to Table \ref{table:cal} in Appendix E for ECE, NLL and BS). Regarding performance, we note that all models perform equally well for ID data, except for cSGHMC, which shows comparatively lower performance. Conversely, under distributional shift, cSGHMC outperforms other models. Figure \ref{Fig:KIT} (e-h) presents histograms of predictive uncertainty on corrupted images. Notably, cSGHMC achieves the highest predictive entropy, followed by DE and MAP. Surprisingly, MCD exhibits even less uncertainty than MAP on shifted data, indicating a higher confidence in incorrect predictions.
%In terms of performance, we observe that for ID, all models perform equally well except cSGHMC that shows worse performance. For distributional shift, we see that cSGHMC is superior. Figure \ref{Fig:KIT} (e-h) reports histograms of predictive entropy on distributionally shifted images. cSGHMC achieves the highest predictive entropy followed by DE and MAP. MCD exhibits even less uncertainty than MAP on shifted data, which means it gives higher confidence incorrect predictions. 

\subsection{Diversity of Model Ensembles}
\label{div_ens}
As we mentioned earlier we hypothesize that models capable of capturing multiple modes in the posterior distribution, produce network samples which are not only accurate but also diverse in their predictions. %When testing on distributionally shifted data,
This diversity in predictions becomes particularly valuable when testing on distributionally shifted data, as each model can provide different predictions over the label space, leading to more reasonable and reliable uncertainty estimates. %In addition to the empirical results of augmented diversity exhibited in Sections 3.2 and 3.3, 
To further support our hypothesis, we calculate the pairwise correlation of softmax outputs for each pair of network samples collected in each method. Figure \ref{cor_matrix} illustrates the average pairwise correlation across six network samples of each method and within different datasets. Among the three methods, cSGHMC exhibits lower correlation among its sample networks compared to the other two. This finding aligns with our earlier results, wherein we observed that cSGHMC consistently provided higher uncertainty estimates in the face of various distribution shifts. Secondly, substantial correlations are evident among all samples of DE across diverse datasets. As demonstrated by our empirical results in the previous section, this lack of diversity translates into an unreliable uncertainty estimation when the model encounters different distributional shifts. As for MCD, the correlations between samples are influenced by the dropout rate. Higher dropout rates result in lower correlations, increased diversity and, improved uncertainty estimation. Nevertheless, it's crucial to highlight that increasing the dropout rate can degrade overall performance. Therefore, a careful consideration is necessary when determining the appropriate dropout rate for each dataset. 
%In our setting, we increased the dropout rate to a level where performance degradation is minimal or the adverse effects are not significant. %it does not degrade the performance or the harm won't be significant and report the results for two levels of dropout rate $p=0.05, 0.1$ for AMOS and KITS and $p=0.05,0.2$ for Hippocampus.
We report the results for two dropout rate levels, namely, $p=0.05$ and $0.1$ for AMOS and KITS, and $p=0.05$ and $0.2$ for Hippocampus. We refer to Table \ref{table:OOD} in Appendix D for the reported Dice scores for MCD across various dropout rates for different datasets.

\section{Discussion} 
%\begin{figure}[t]
%     \centering
%     \begin{subfigure}[b]{0.3\textwidth}
%         \centering
%         \includegraphics[width=\textwidth]{New_Images/hippo_sanity.png}
         %\caption{CIFAR-10}
         %\label{in_vivo_10_dice}
%     \end{subfigure}
%     \hfill
%     \begin{subfigure}[b]{0.3\textwidth}
%         \centering
%         \includegraphics[width=\textwidth]
%         {New_Images/kit_sanity.png}
         %\caption{CIFAR-100}
         %\label{in_vivo_2_dice}
%     \end{subfigure}
%     \hfill
%     \begin{subfigure}[b]{0.3\textwidth}
%         \centering
%         \includegraphics[width=\textwidth]{New_Images/amos_sanity.png}
         %\caption{ImageNet-10}
         %\label{dice_D}
%     \end{subfigure}
%     \\
%     \centering
%     \begin{subfigure}[b]{0.32\textwidth}
%         \centering
%         \includegraphics[width=\textwidth]{New_Images/dens_hippo.png}
%         \caption{Hippocampus}
%         \label{mds:hippo}
%     \end{subfigure}
%     \hfill
%     \begin{subfigure}[b]{0.32\textwidth}
%         \centering
%         \includegraphics[width=\textwidth]
%         {New_Images/dens_kit.png}
%         \caption{KITS}
%         \label{mds:kit}
%     \end{subfigure}
%     \hfill
%     \begin{subfigure}[b]{0.32\textwidth}
%         \centering
%         \includegraphics[width=\textwidth]{New_Images/dens_amos.png}
%         \caption{AMOS}
%         \label{mds:amos}
%     \end{subfigure}
%\caption{Diversity in the weight space}
%\label{Div:WS}
%\end{figure}

\begin{figure}[!t]
\centering
\begin{subfigure}{1.0\textwidth}
    \includegraphics[width=\textwidth]{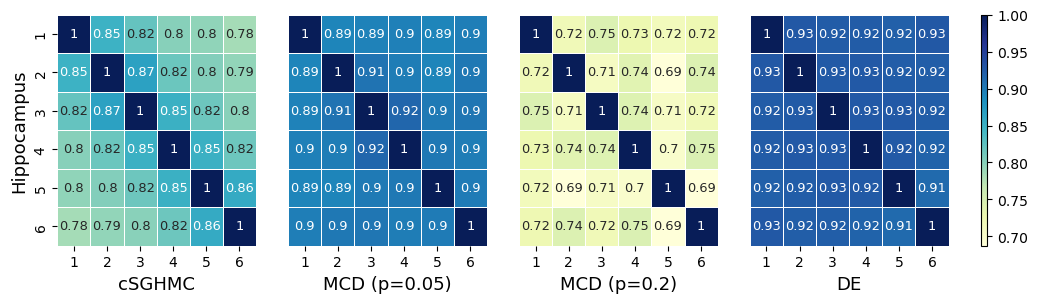}
    %\caption{CIFAR-10}
    %\label{fig:acc_cif10}
\end{subfigure}
\\
\begin{subfigure}{1.0\textwidth}
    \includegraphics[width=\textwidth]{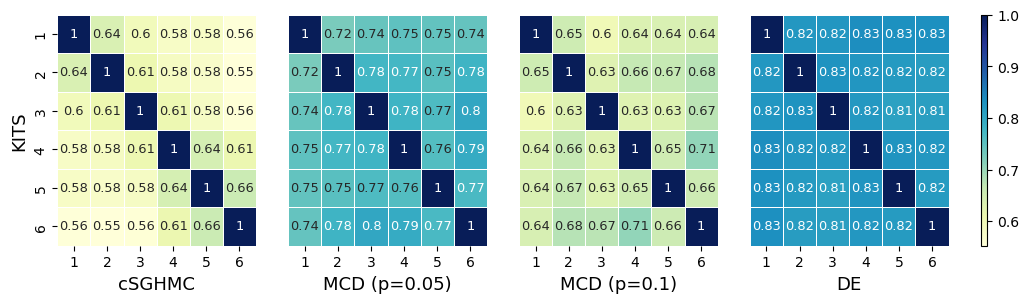}
    %\caption{CIFAR-10}
    %\label{fig:nll_cif10}
\end{subfigure}
\\
\centering
\begin{subfigure}{1.0\textwidth}
    \includegraphics[width=\textwidth]{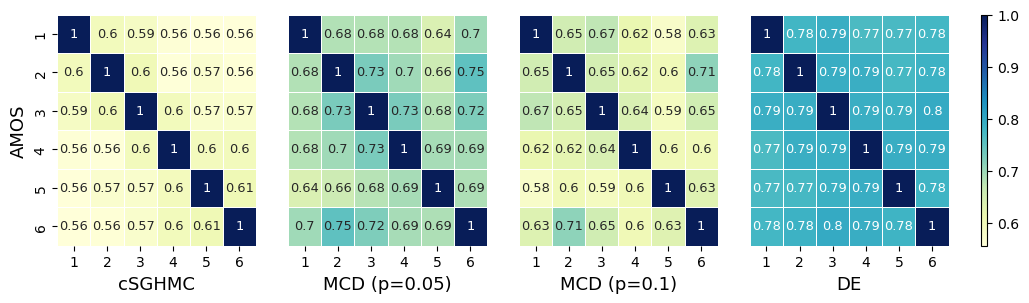}
    %\caption{CIFAR-100}
    %\label{fig:acc_cif100}
\end{subfigure}\hskip1ex
\caption{Pairwise correlation of softmax outputs between any two model samples in the posterior distribution for cSGHMC and in the predictive space for MCD and DE. For cSGHMC, samples from the last 6 cycles were chosen. For MCD and DE, 6 models were randomly selected.}
\label{cor_matrix}
\end{figure}
We conducted a comprehensive evaluation of commonly used methods for quantifying predictive uncertainty, focusing specifically on their applicability in medical domain, under dataset shift. Our take-home messages are as follows:
\begin{itemize}
\item[$\bullet$] 
While determining a method that consistently outperforms others proves challenging, our results underscore the significance of techniques capable of exploring and characterizing \textit{multiple modes} in the posterior distribution, as they offer more \textit{reliable} predictive uncertainty.
%Although determining a method that consistently performs better seems challenging, the results highlight that the methods that are capable of exploring and characterizing \textit{multiple modes} in the posterior distribution, offer more \textit{reliable} predictive uncertainty.
Our empirical findings demonstrate that cSGHMC indicates more consistent behaviour by generating high entropy when faced with various distribution shifts. In contrast, methods like MCD and DE show less robust behavior across different distributional shifts, with MCD showing high uncertainty in covariate shift but performs poorly in modality and corruption shifts. On the other hand, DE tends to underperform in covariate shift while exhibiting higher uncertainty in the other two types of shifts.

\item[$\bullet$]
Confirming the findings in \citet{trust}, we observe that calibration within the ID does not necessarily guarantee calibration under distributional shift in real-world datasets. Specifically, for covariate and modality shifts the models are well-calibrated for ID, but as the intensity of the shift increases, they become miscalibrated. Notably, for corruption shift, cSGHMC remains well-calibrated for both the ID and distribution shift, while other methods demonstrate miscalibration in distribution shift. 

\item[$\bullet$]
Despite the established belief about DE that training from scratch converges to different basin in the posterior distribution \citep{GPB}, our findings reveal that DE encounters challenges in achieving model diversity (see Figure \ref{cor_matrix} last column). This limitation hampers its ability to offer reliable uncertainty estimates across diverse distributional shifts (see Figures \ref{fig:dehip} and \ref{fig:rot_enc_unc}). This is while in DE, we are obligated to train multiple models from scratch, imposing a significant computational burden.

%Moreover, it provides more calibrated uncertainty than MCD and DE. DE and MCD did not indicate consistent behaviour across different distributional shifts. While MCD provides high uncertainty on covariate shift, it performs poorly on modality and corruption shifts. Likely, DE tends to perform poorly in covariate shift, while it exhibits higher uncertainty in the other two types of shifts. Moreover, despite the established belief about DE that training from scratch converges to different basin in the posterior distribution \citep{GPB}, our findings reveal that DE encounters challenges in achieving model diversity. This limitation results in an inability to offer reliable uncertainty estimates across diverse distributional shifts.

%These results suggest that the methods that are capable of exploring and characterizing \textit{multiple modes} in the posterior distribution, provide more \textit{reliable} predictive uncertainty. To further analyze this, following \cite{cSGMCMC}, we employ Multidimensional
%Scaling (MDS) to visualize the 20 samples we collected using cSGHMC in three datasets in Figure \ref{Div:WS}. We see that the samples of cSGHMC form different clusters, which means they are from different modes in weight space. Each mode characterizes a meaningfully different representation of data. When testing on the shifted data, each mode can provide different predictions over the label space, leading to more reasonable uncertainty estimates. 

\item[$\bullet$]
%For cSGHMC, we observe that the degradation in performance corresponds to a high level of uncertainty when dealing with shifted data.
In the case of cSGHMC, we observe that the decline in performance corresponds with an increased level of uncertainty when encountering shifted data across different distribution shifts. This realization empowers practitioners to recognize that, when attempting to deploy the model, if the dataset has a distribution different from the training data. They can do so by establishing a threshold on the uncertainty level, when the uncertainty surpasses this threshold, practitioners can make informed decisions to refrain from deploying the model, avoiding potential inaccuracies and erroneous decisions. 

\end{itemize}
The effective deployment of deep learning models in medical applications demands not only high predictive accuracy but also a reliable quantification of uncertainty. This becomes especially paramount in scenarios where model predictions influence critical decision-making processes. By shedding light on methods that excel in producing \textit{well-calibrated uncertainty}, our work aims to enhance the confidence of practitioners in deploying deep learning models autonomously, paving the way for their broader and more impactful use in medical settings. The insights gained from our study can potentially contribute to advancing the field's understanding of uncertainty quantification, fostering the development of more reliable and robust uncertainty quantification methods for medical applications. 

% github repo for hasan's thesis:
%https://github.com/tasnimul-unipotsdam/ms_thesis

% References
\bibliography{ref}

\newpage
\section*{Appendix A}
\label{app:A}

\begin{figure}[h]
\centering
\begin{minipage}[c]{1.0\linewidth}
    \begin{subfigure}{1.1\linewidth}
        \includegraphics[width=\linewidth]{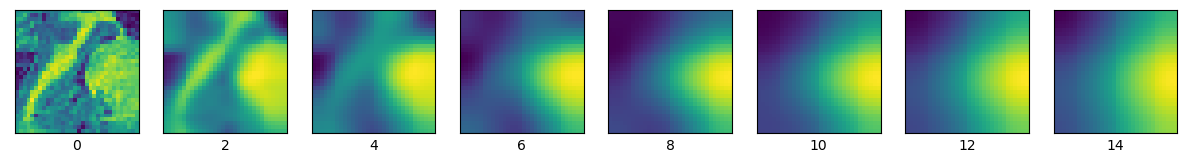}
    \end{subfigure}
    \\
    \begin{subfigure}{1.1\linewidth}
        \includegraphics[width=\linewidth]{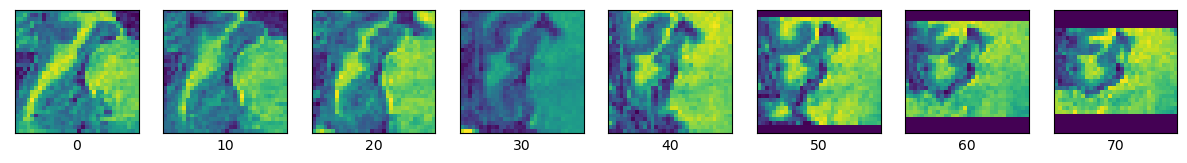}
    \end{subfigure}

\caption{Examples of Hippocampus images corrupted by: Gaussian blur (first row), at intensities of 0 (uncorrupted
image) through 14 (maximum corruption), and Rotation (second row) at angels 0 (uncorrupted image) through 70.}
\label{Fig:cor_hippo}
\end{minipage}
\end{figure}

%\begin{figure}[!h]
%    \includegraphics[width=1.1\textwidth]{New_Images/hipo_blur_dist.png}
%\end{figure}

%\begin{figure}[!h]
%    \includegraphics[width=1.1\textwidth]{New_Images/hipo_rot_dist.png}
%\end{figure}

\begin{figure}[h]
\centering
\begin{minipage}[c]{1.0\linewidth}
    \begin{subfigure}{1.1\linewidth}
        \includegraphics[width=\linewidth]{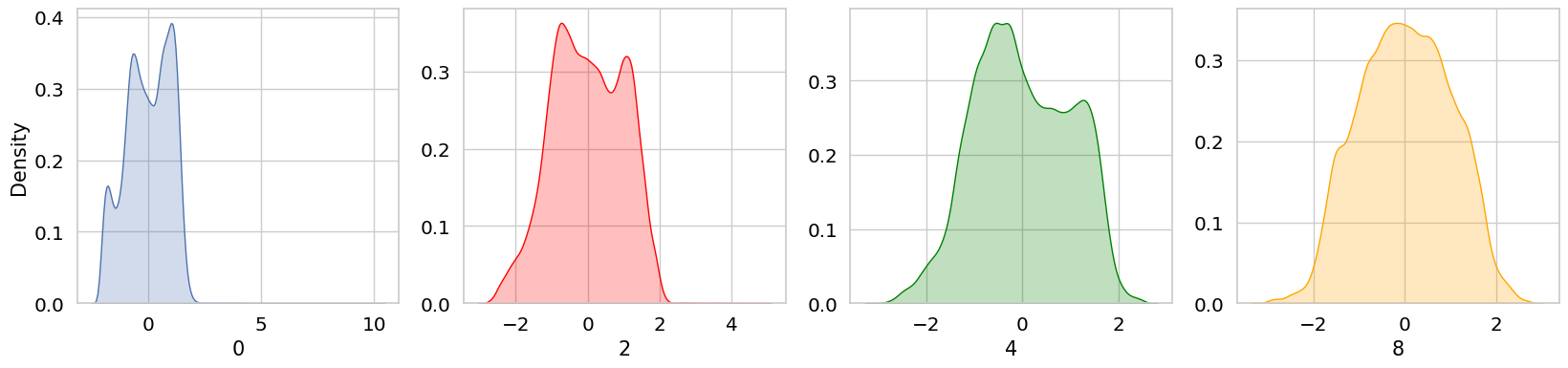}
        \label{blur_dist}
        %\caption{\normalsize ex vivo 14}
    \end{subfigure}
    \\
    \begin{subfigure}{1.1\linewidth}
        \includegraphics[width=\linewidth]{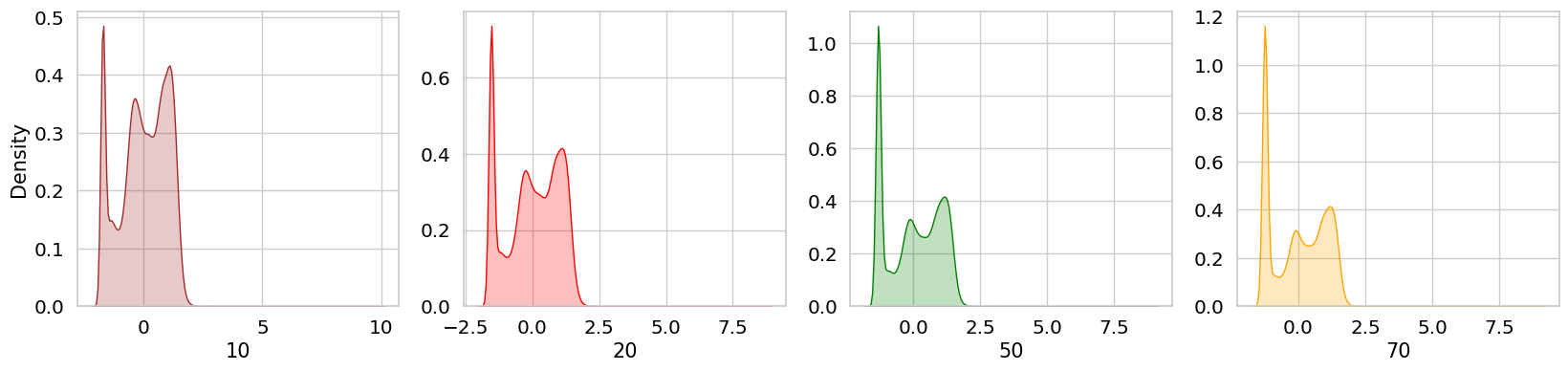}
        \label{rot_dist}
        %\caption{\normalsize ground truth}
    \end{subfigure}

\caption{Examples of KDE plots for Hippocampus corrupted images. First row: KDE plots of blurred images at intensities 0, 2, 4 and 8. Second row: KDE plots of rotated images at angels 10, 20, 50 and 70.}
\label{Fig:kde_hippo}
\end{minipage}
\end{figure}

\newpage
\section*{Appendix B}
\label{app:B}
%\begin{figure}[!h]
%\centering
%    \includegraphics[width=0.48\textwidth]{New_Images/AMOS_img.png}
%\end{figure}

%\begin{figure}[!h]
%\centering
%    \includegraphics[width=0.52\textwidth]{New_Images/AMOS_dist.png}
%\end{figure}
\begin{figure}[!h]
\centering
    \begin{subfigure}{0.48\linewidth}
        \includegraphics[width=\linewidth]{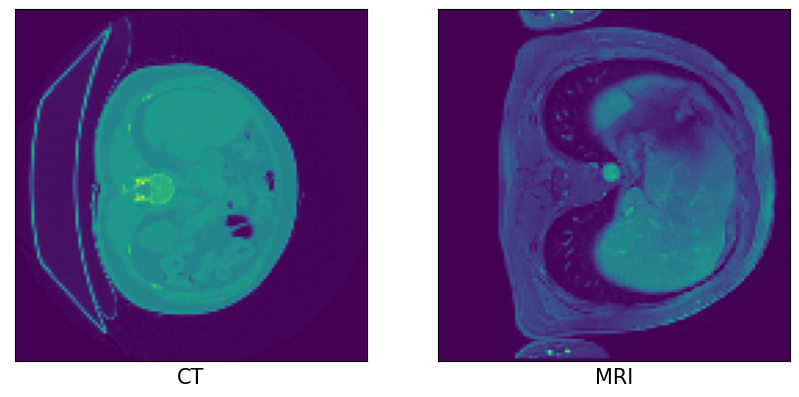}
        %\caption{\normalsize ex vivo 14}
    \end{subfigure}
    \\
    \begin{subfigure}{0.52\linewidth}
        \includegraphics[width=\linewidth]{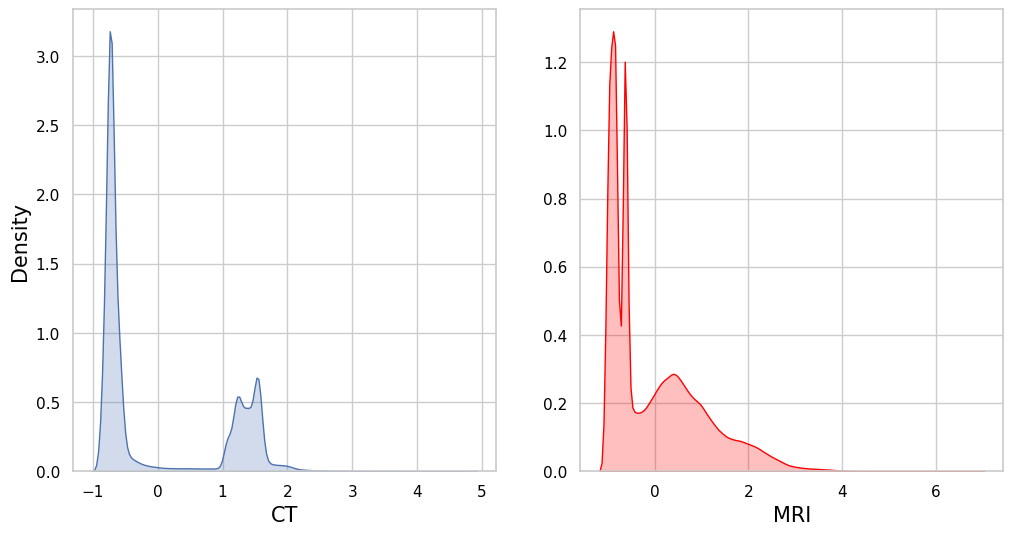}
        %\caption{\normalsize ground truth}
    \end{subfigure}

\caption{First row: Examples of AMOS images, CT and MRI scans. Second row: Their corresponding KDE plots.}
\label{Fig:amos_kde}
\end{figure}

\section*{Appendix C}
\label{app:C}
%\begin{figure}[!h]
%\centering
%    \includegraphics[width=0.8\textwidth]{New_Images/kits_corr.png}
%\end{figure}
%\begin{figure}[!h]
%\centering
%    \includegraphics[width=0.6\textwidth]{New_Images/KITS_dist.png}
%\end{figure}

\begin{figure}[!h]
\centering
    \begin{subfigure}{0.8\linewidth}
        \includegraphics[width=\linewidth]{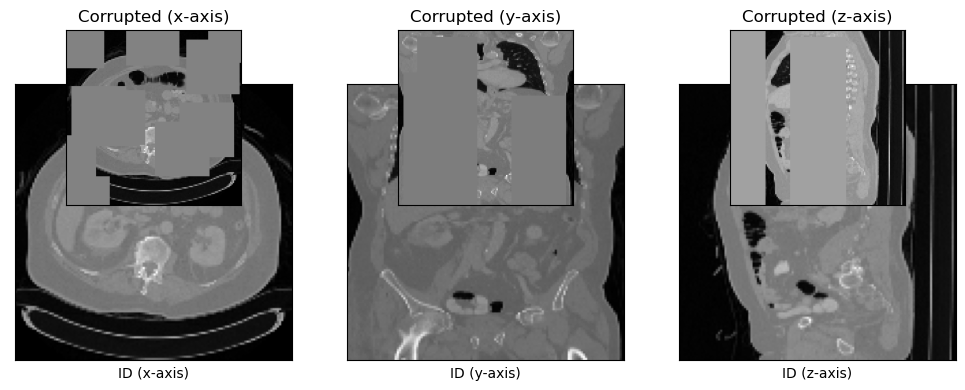}
        %\caption{\normalsize ex vivo 14}
    \end{subfigure}
    \\
    \begin{subfigure}{0.6\linewidth}
        \includegraphics[width=\linewidth]{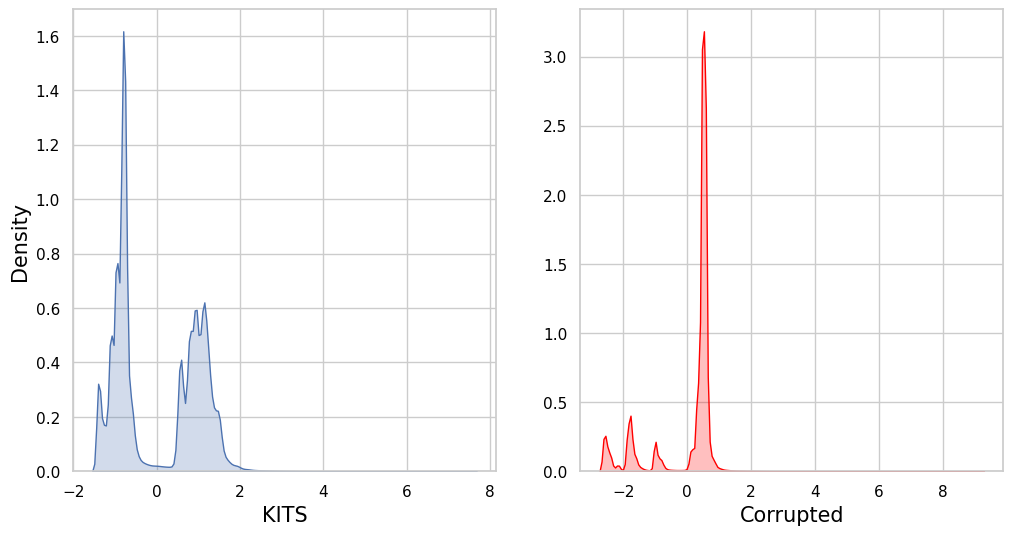}
        %\caption{\normalsize ground truth}
    \end{subfigure}

\caption{First row: Examples of KITS scans and their corrupted versions. Second row: KDE plots of original images and corrupted images.}
\label{Fig:kits_kde}
\end{figure}

\newpage
\section*{Appendix D}
\label{app:D}
\paragraph{Hippocampus} For Hippocampus dataset, we train cSGHMC for $1200$ epochs with an initial learning rate of $0.1$ and a cycle length of $50$ epochs. The Langevin noise is injected from epoch $10$ and the first $200$ epochs are discarded as burun-in time. We train other methods for $1000$ epochs, with a learning rate of $0.00001$ and dropout rate of $0.2$. For all models a weight decay of $0.0001$ is used. 

\paragraph{AMOS} For AMOS dataset, we train cSGHMC for $2500$ epochs with an initial learning rate of $0.1$ and cycle length of $100$. The Langevin noise is injected from epoch $40$ and the first $500$ epochs are discarded as burn-in time. The other approaches are trained for $2000$ epochs with learning rate of $0.0001$ and dropout rate of $0.05$ during training. For MCD, the dropout rate is set to $0.05$ at evaluation time. For all methods we use weight decay equal to $0.00001$.

\paragraph{KITS} For KITS dataset, we train cSGHMC for $2500$ epochs with an initial learning rate of $0.1$ and cycle length of $50$. We inject the Langevin noise from epoch $45$ and discard the first $500$ epochs as burn-in time. We train other methods for $2000$ epochs with learning rate of $0.0001$ and dropout rate of $0.2$. For MCD, the dropout at test time is set to $0.05$ and we use weight decay equal to $0.00001$ for all methods.

For choosing dropout rate at evaluation time for MCD, we traverse over dropout rates $[0.01,0.05,0.1,0.2,0.3]$ for all datasets. We  select the dropout rate that gives the best Dice score on ID test set. For the AMOS dataset, we select a dropout rate of $0.05$. This choice improves calibration and uncertainty estimation, as shown in Table \ref{table:cal}, albeit at the expense of a lower dice score. The performance for different dropout rates are reported in Table \ref{table:OOD}. The results are rounded in three digits. 

\begin{table}[!h]
\caption{Dice score for MCD across different datasets for various dropout rates}  %Results obtained from our method are underlined and the best results are indicated in bold.} 
\medskip
\centering
\begin{tabular}{l| c c c c c}
\hline
\textbf{Data/ dropout rate}& $0.01$ & $0.05$ & $0.1$ & $0.2$ & $0.3$ \\
\hline
Hippocampus & $0.888$ & $0.891$ & $0.892$ & $\textbf{0.892}$ & $0.877$ \\
\hline
AMOS & $\textbf{0.872}$ & $0.863$ & $0.811$& - & -   \\
\hline
KITS & $0.936$  & $\textbf{0.936}$ & $0.935$ & - & - \\
\end{tabular}
\label{table:OOD}
\end{table}

\begin{table}[!h]
\caption{Dice score for cSGHMC and DE across different datasets}  %Results obtained from our method are underlined and the best results are indicated in bold.} 
\medskip
\centering
\begin{tabular}{l| c c }
\hline
\textbf{Data/ Method}& cSGHMC & DE  \\
\hline
Hippocampus & 0.87 & 0.89 \\
\hline
AMOS & 0.83 &  0.88   \\
\hline
KITS & 0.91 & 0.93  \\
\end{tabular}
\label{table:OOD_OM}
\end{table}

%\begin{table}[!htb]
%\begin{minipage}{.5\linewidth}
%    \centering

%    \caption{Variation in Dice Score for MCD at Different Dropout Rates}
%    \label{tab:first_table}

%    \medskip

%\begin{tabular}{ l c c c} 
%\toprule
%\makecell{Data/ dropout rate}  & $0.05$ & $0.1$ & $0.2$ \\   
%\midrule
%Hippocampus & $0.891$ & $0.892$ & $\textbf{0.892}$  \\
%AMOS & $\textbf{0.872}$ & $0.811$& -   \\
%KITS & $0.936$   & $0.935$ & -  \\
%\bottomrule
%\end{tabular}
%\end{minipage}\hfill
%\begin{minipage}{.5\linewidth}
%    \centering

%    \caption{Second Table}
%    \label{tab:second_table}

%    \medskip

%\begin{tabular}{ l| c c } 
%    \toprule
%    \makecell{Data/ Method}  & cSGHMC & DE  \\   
%    \midrule
%    Hippocampus & 0.87 & 0.89 \\
%    AMOS & 0.83 &  0.88   \\
%    KITS & 0.91 & 0.93  \\
%    \bottomrule
%\end{tabular}
%\end{minipage}
%\end{table}

\newpage
\section*{Appendix E}
\label{app:E}
Table \ref{table:cal} presents various calibration metrics on the AMOS and KITS datasets, considering both ID and distribution shift. While the ranking of methods varies for ID and shifted data across different metrics and datasets, the following results are observed. For shifted data in the AMOS dataset, DE and MCD perform slightly better than cSGHMC, with DE exhibiting superior performance. Conversely, for shifted data in KITS, cSGHMC outperforms other methods across all metrics by a significant margin.

In the case of ID on the AMOS dataset, cSGHMC excels in terms of ECE with a substantial margin. Regarding BS, DE performs slightly better than the other two methods, and for NLL, MCD outperforms the other two. On the KITS dataset for ID, cSGHMC and MCD indicate lower NLL and better ECE. In terms of BS, DE performs slightly better than the other methods.

While drawing a definitive conclusion on which method exhibits superior calibration across various datasets and shifts proves challenging, on average, we observe that cSGHMC outperforms the other two methods. cSGHMC demonstrates better performance in three metrics for distribution shift and two metrics for ID when compared to MCD, which excels in two ID metrics. Additionally, compared to DE, which shows better performance in three distribution shift metrics and two ID metrics (albeit slightly), cSGHMC performs at a comparable level.
%Although, drawing a conclusion on which method is better calibrated across different datasets and shifts is difficult, but in average we observe that cSGHMC performs better than the other two methods (with better performance in 3 metrics in distribution shift and 2 metrics in ID) compared to MCD (with superior performance in 2 metrics in ID) and DE (with better performance in 3 distribution shift metrics and 2 metrics in ID which are slightly better).   

%For ID, in AMOS dataset cSGHMC outperforms other methods in terms of ECE with significant margin, while in terms of BS, DE performs slightly better than the other two methods and in terms of NLL, MCD outperforms the other two methods. For ID data in KITS, cSGHMC and MCD indicate lower NLL and better ECE. In terms of Brier score, DE performs slightly better than the other methods.       
%Moreover, we observe that MCD with higher dropout rate (p=$0.05$) is more calibrated at the cost of lower Dice score. 

\begin{table}[!h]
\caption{Assessing calibration of different methods for AMOS and KITS datasets in terms of different metrics. The best results are indicated in bold.}  %Results obtained from our method are underlined and the best results are indicated in bold.} 
\medskip
\centering
\begin{tabular}{l  c c c |  c c c}
\hline
 & \multicolumn{6}{c}{AMOS}\\
 \hline
  & \multicolumn{3}{c|}{\textbf{In Distribution}} & \multicolumn{3}{c}{\textbf{Distribution Shift}} \\
\hline
\textbf{Method} & BS $\downarrow$ & ECE $\downarrow$ & NLL $\downarrow$ & BS $\downarrow$ & ECE $\downarrow$ & NLL $\downarrow$  \\
\hline
MAP &$1.35 *10^{-2} $& $3.7 *10^{-3}$ & $4.4 *10^{-2}$ & $1.37 *10^{-1}$ & $6.4*10^{-2}$ & $7.2 *10^{-1}$ \\ 
MCD (p=0.01) & $1.31 * 10^{-2}$  & $3.1*10^{-3}$ & $4.1*10^{-2}$ & $1.34*10^{-1}$ & $6.2*10^{-2}$ & $6.7*10^{-1}$ \\
MCD (p=0.05) & $1.26*10^{-2}$ & $91.1*10^{-3}$ & $\mathbf{2.8}*10^{-2}$ & $1.29*10^{-1}$ & $5.8 *10^{-2}$ & $5.7*10^{-1}$ \\
DE & $\mathbf{1.25} *10^{-2}$ & $2.3 *10^{-3}$ & $3.7*10^{-2}$ & $\mathbf{1.27} *10^{-1}$  & $\mathbf{5.6}*10^{-2}$ & $\mathbf{5.5}*10^{-1}$  \\
cSGHMC & $1.5*10^{-2}$ & $\mathbf{1.6}*10^{-3}$  & $4.1*10^{-2}$  &  $1.29 *10^{-1}$ & $5.9*10^{-2}$  & $5.8*10^{-1}$  \\
\hline
 & \multicolumn{6}{c}{KITS}\\
 \hline
MAP  &$4.16 *10^{-3} $& $5.2*10^{-4}$ & $2.38*10^{-2}$  & $8.27*10^{-2}$ & $4.1*10^{-2}$ & $19*10^{-2} $\\ 
MCD  & $4.02 * 10^{-3}$  & $\mathbf{1.0}*10^{-4}$ & $1.77*10^{-2}$  & $7.16*10^{-2} $ & $2.9*10^{-2} $ & $14*10^{-2} $ \\
DE & $\mathbf{4.0} *10^{-3} $ & $ 3.1*10^{-4}$ & $ 2.26*10^{-2}$ & $ 6.46*10^{-2}$  & $ 2.2*10^{-2}$ & $ 13*10^{-2}$  \\
cSGHMC  & $4.59*10^{-3} $ & $ 1.4*10^{-4}$  & $\mathbf{1.62}*10^{-2} $  &  $\mathbf{1.58}*10^{-2}$ & $\mathbf{0.34}*10^{-2}$  & $\mathbf{5.07}*10^{-2}$  \\ 
\end{tabular}
\label{table:cal}
\end{table}

%\newpage
%\section*{Appendix F}
%\label{app:F}
%\begin{figure}[!h]
%\centering
%\begin{minipage}[c]{1.0\linewidth}
%    \begin{subfigure}{1.1\linewidth}
%        \includegraphics[width=\linewidth]{New_Images/Amos/amos_cor_diff_mcd.png}
        %\caption{\normalsize ex vivo 14}
%    \end{subfigure}
%    \\
%    \begin{subfigure}{1.1\linewidth}
%        \includegraphics[width=\linewidth]{New_Images/Kits/cor_kits_diff_mcd.png}
        %\caption{\normalsize ground truth}
%    \end{subfigure}
%\caption{Pairwise correlation of softmax outputs between any two samples of cSGHMC and MCD samples. The correlation decreases as we vary drop out rate.}
%\label{fig:cor_mat}
%\end{minipage}
%\end{figure}

\newpage
\section*{Appendix G}
\label{app:G}
\begin{figure}[!h]
\centering
\begin{minipage}[c]{1.0\linewidth}
    \begin{subfigure}{1.1\linewidth}
        \includegraphics[width=\linewidth]{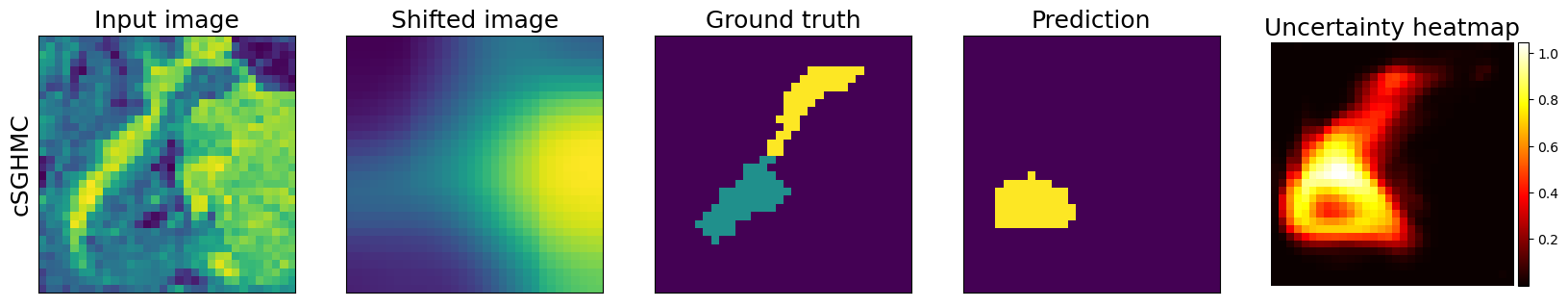}
        %\caption{\normalsize ex vivo 14}
    \end{subfigure}
    \\
    \begin{subfigure}{1.1\linewidth}
        \includegraphics[width=\linewidth]{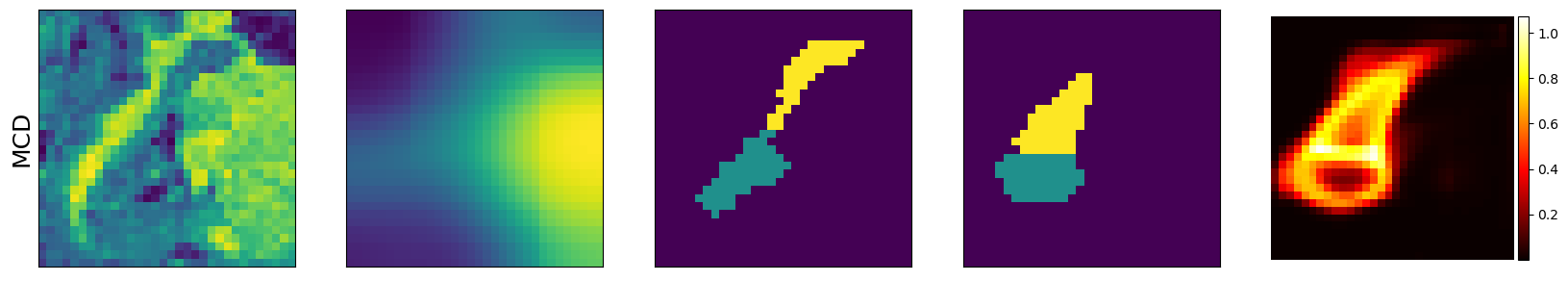}
        %\caption{\normalsize ground truth}
    \end{subfigure}
    \\
    \begin{subfigure}{1.1\linewidth}
        \includegraphics[width=\linewidth]{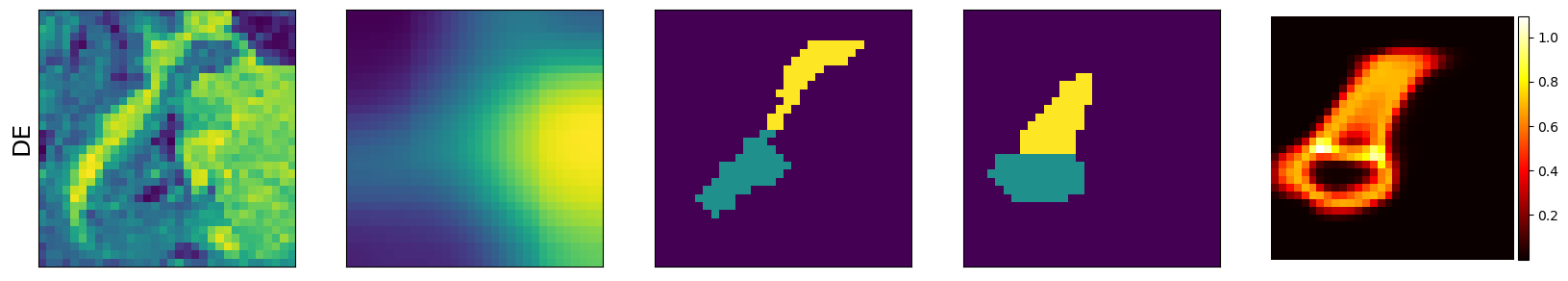}
        %\caption{\normalsize prediction}
    \end{subfigure}
    \\
    \begin{subfigure}{1.1\linewidth}
        \includegraphics[width=\linewidth]{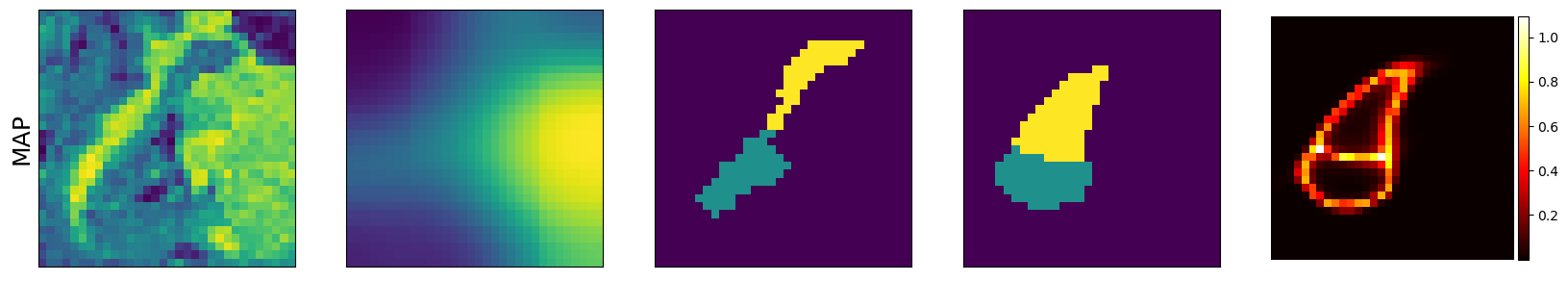}
        %\caption{\normalsize uncertainty}
    \end{subfigure}
\end{minipage}
\caption{Visualization of images from the Hippocampus dataset, including original data, data shifted by Gaussian blur at intensity 6, ground truth, predictions (on shifted data), and uncertainty maps for each method. MAP and DE assign lower uncertainty, as demonstrated by the histogram plots in Figure \ref{Fig:Hipo} in the main paper.}
\label{Hippo_blur}
\end{figure}

\begin{figure*}[t]
\centering
\begin{minipage}[c]{1.0\linewidth}
    \begin{subfigure}{1.1\linewidth}
        \includegraphics[width=\linewidth]{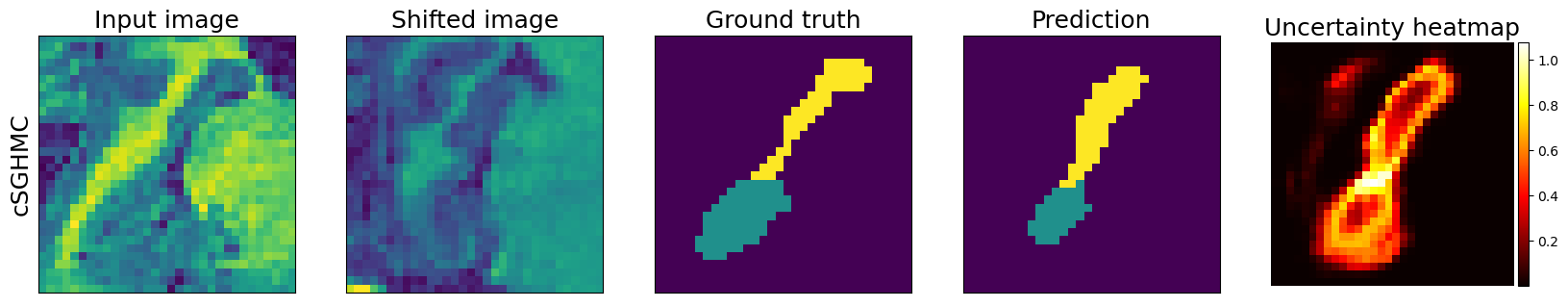}
        %\caption{\normalsize ex vivo 14}
    \end{subfigure}
    \\
    \begin{subfigure}{1.1\linewidth}
        \includegraphics[width=\linewidth]{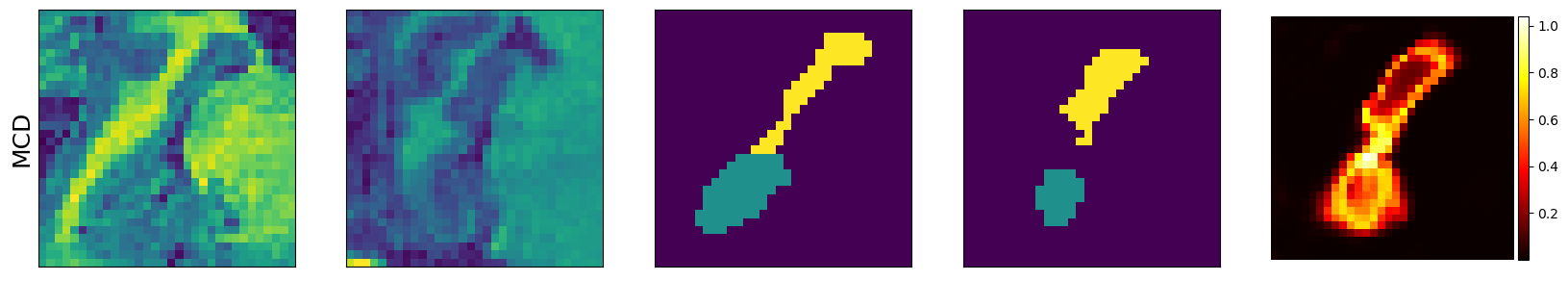}
        %\caption{\normalsize ground truth}
    \end{subfigure}
    \\
    \begin{subfigure}{1.1\linewidth}
        \includegraphics[width=\linewidth]{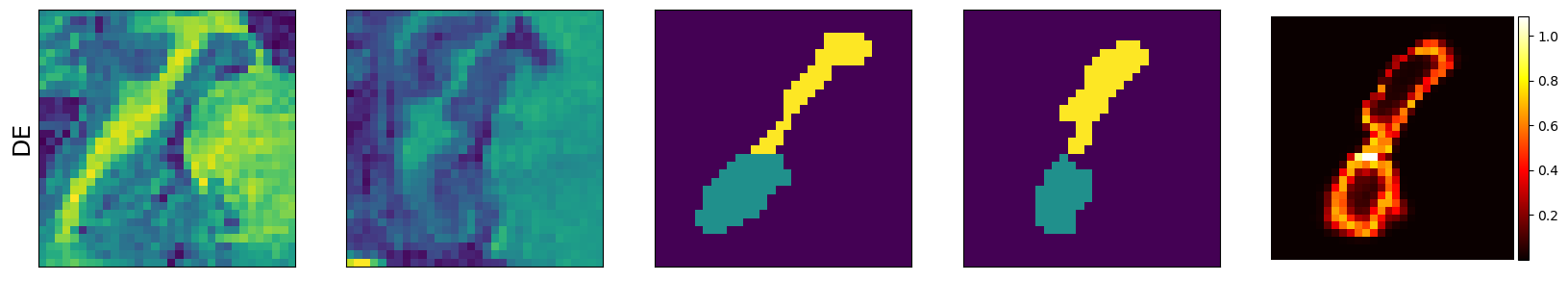}
        %\caption{\normalsize prediction}
    \end{subfigure}
    \\
    \begin{subfigure}{1.1\linewidth}
        \includegraphics[width=\linewidth]{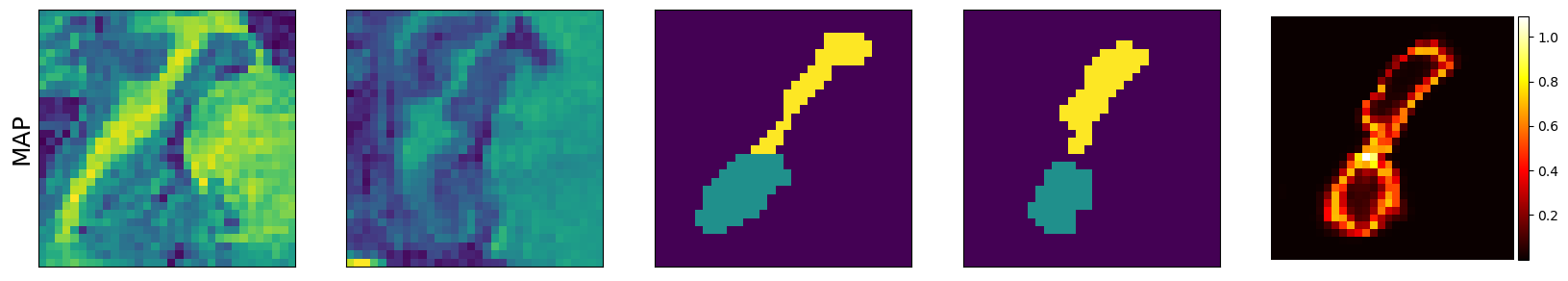}
        %\caption{\normalsize uncertainty}
    \end{subfigure}
\end{minipage}
\caption{Visualization of images from the Hippocampus dataset, including original data, data shifted by Rotation at angle 30, ground truth, predictions (on shifted data), and uncertainty maps for each method. MAP and DE assign lower uncertainty, as demonstrated by histogram plots in Figure \ref{Fig:rot}.}
\label{Hippo_rot}
\end{figure*}

\begin{figure*}[t]
\centering
\begin{minipage}[c]{1.0\linewidth}
    \begin{subfigure}{1.1\linewidth}
        \includegraphics[width=\linewidth]{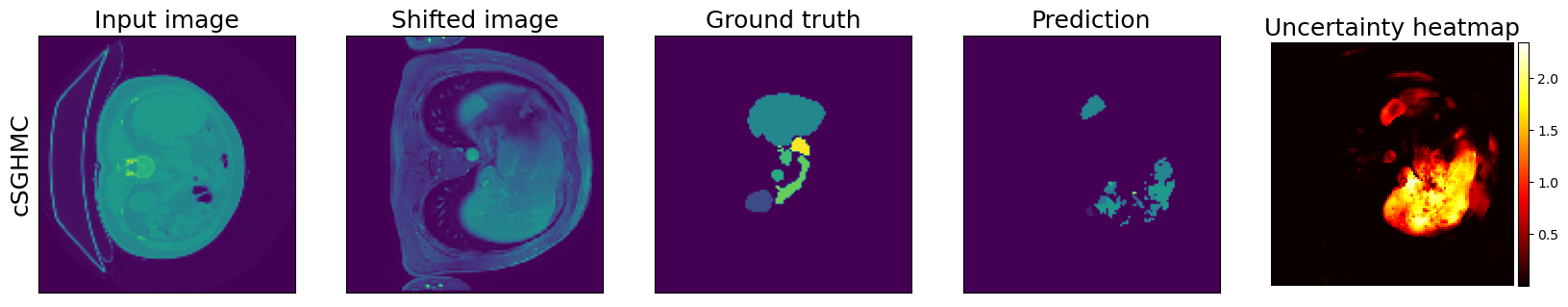}
        %\caption{\normalsize ex vivo 14}
    \end{subfigure}
    \\
    \begin{subfigure}{1.1\linewidth}
        \includegraphics[width=\linewidth]{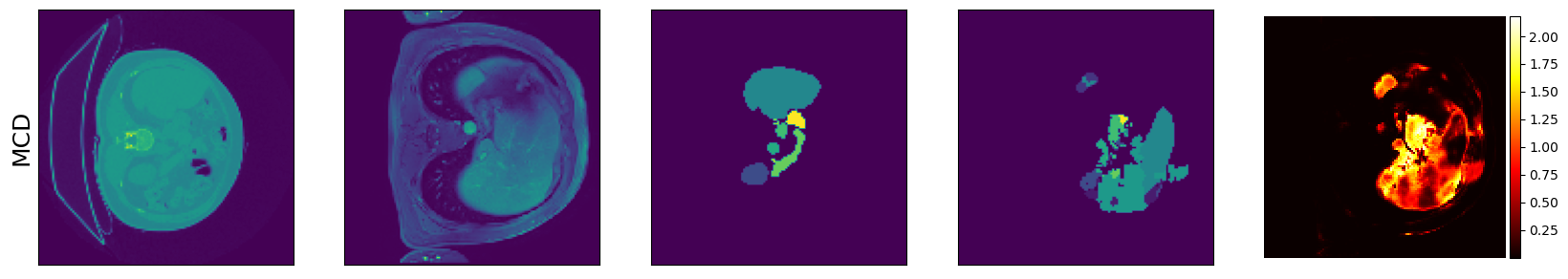}
        %\caption{\normalsize ground truth}
    \end{subfigure}
    \\
    \begin{subfigure}{1.1\linewidth}
        \includegraphics[width=\linewidth]{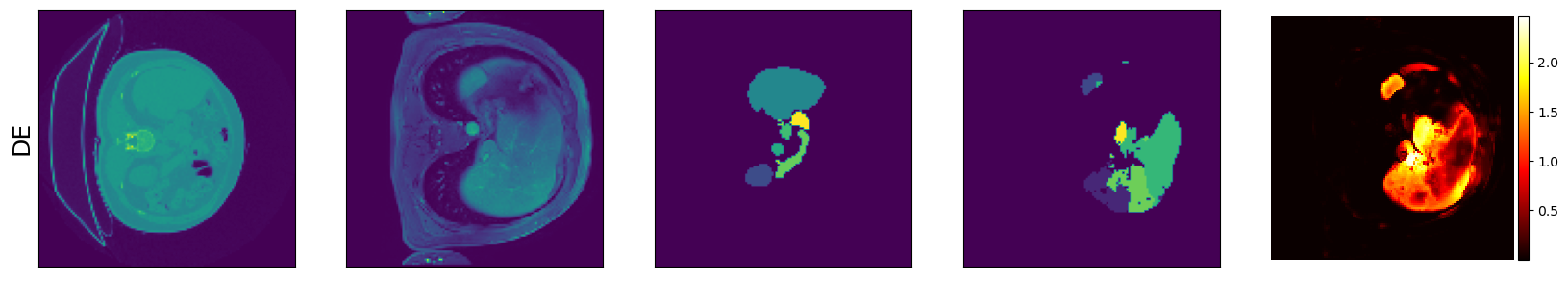}
        %\caption{\normalsize prediction}
    \end{subfigure}
    \\
    \begin{subfigure}{1.1\linewidth}
        \includegraphics[width=\linewidth]{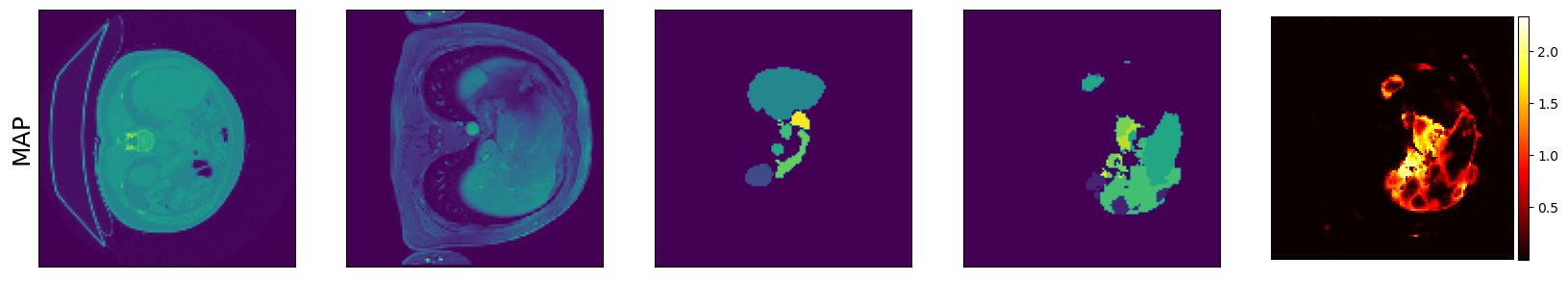}
        %\caption{\normalsize uncertainty}
    \end{subfigure}
\end{minipage}
\caption{Visualization of images from the AMOS dataset including original data (CT), shifted data (MRI), ground truth (for CT), predictions (on MRI) and uncertainty maps for each method. Except MAP, the methods show nearly identical levels of uncertainty.}
\label{Amos:mod}
\end{figure*}

\begin{figure*}[t]
\centering
\begin{minipage}[c]{1.0\linewidth}
    \begin{subfigure}{1.1\linewidth}
        \includegraphics[width=\linewidth]{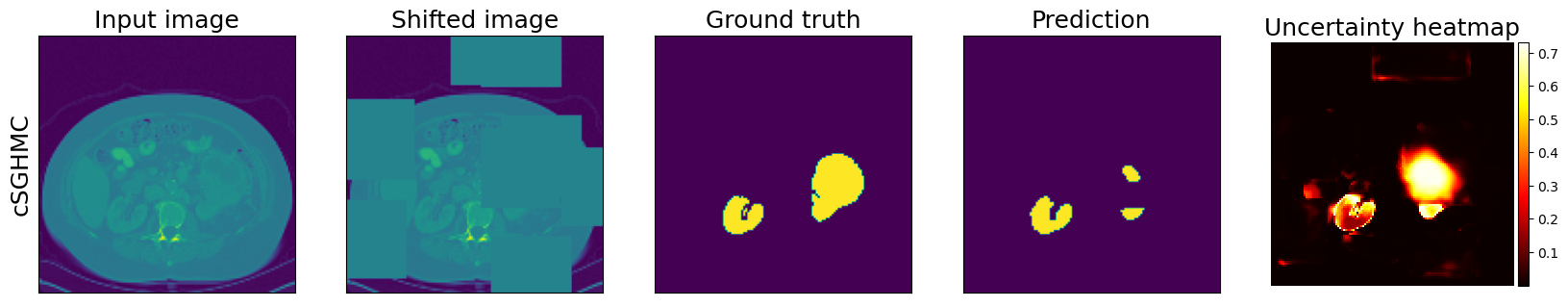}
        %\caption{\normalsize ex vivo 14}
    \end{subfigure}
    \\
    \begin{subfigure}{1.1\linewidth}
        \includegraphics[width=\linewidth]{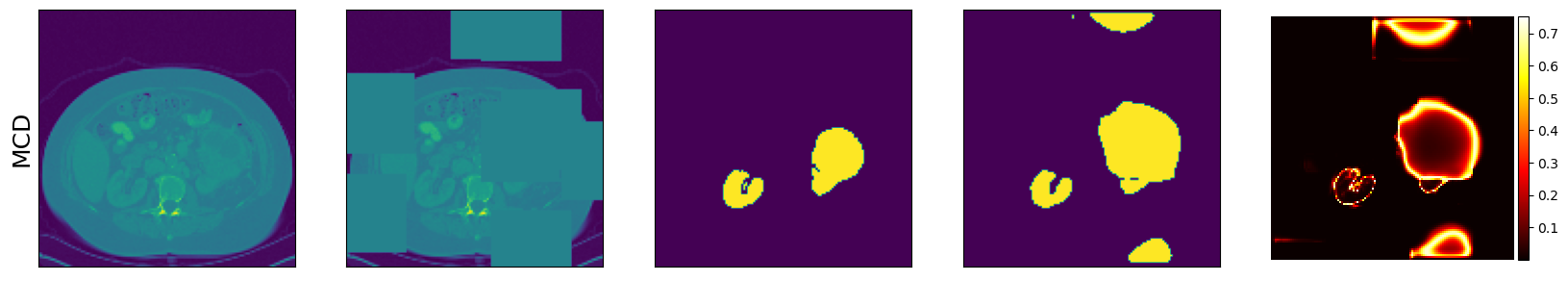}
        %\caption{\normalsize ground truth}
    \end{subfigure}
    \\
    \begin{subfigure}{1.1\linewidth}
        \includegraphics[width=\linewidth]{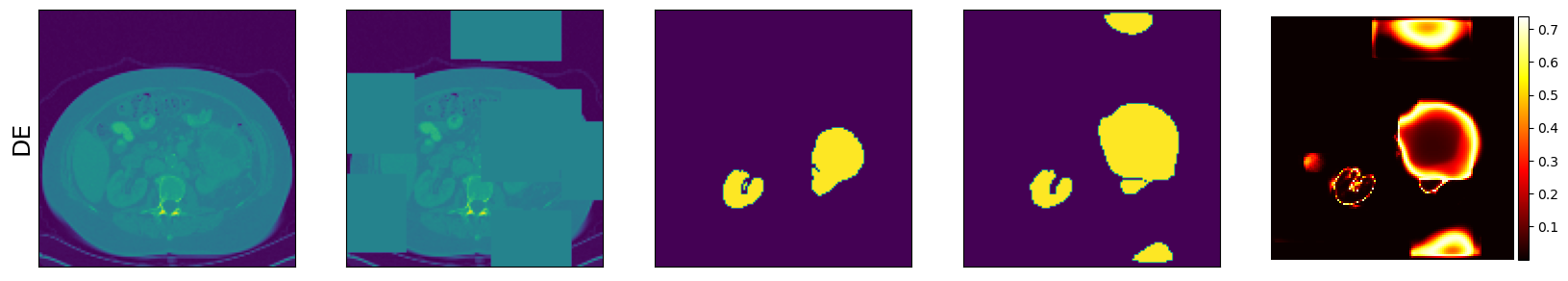}
        %\caption{\normalsize prediction}
    \end{subfigure}
    \\
    \begin{subfigure}{1.1\linewidth}
        \includegraphics[width=\linewidth]{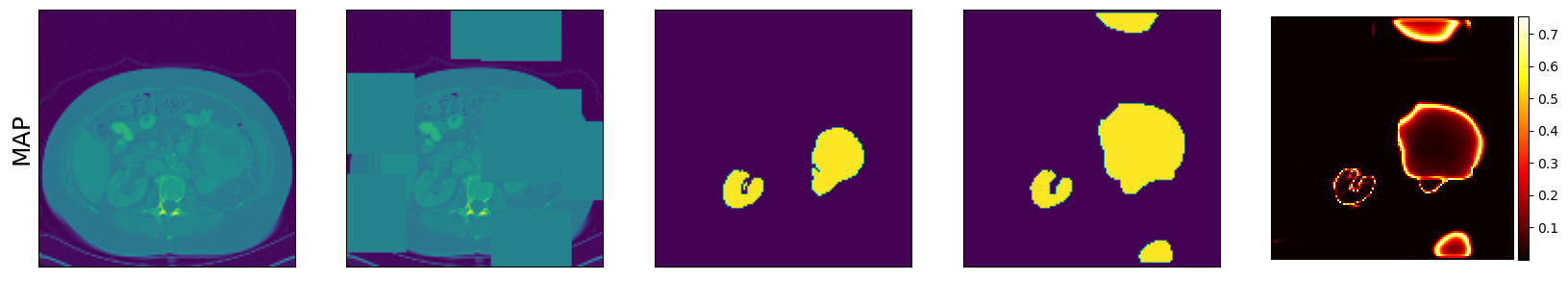}
        %\caption{\normalsize uncertainty}
    \end{subfigure}
\end{minipage}
\caption{Visualization of images from the KITS dataset including original data, corrupted images, ground truth, predictions (on corrupted images) and uncertainty maps for each method. MAP and MCD indicate lower uncertainty, as demonstrated by histogram plots in Figure \ref{Fig:KIT} in the main paper.}
\label{kits}
\end{figure*}

In Figures \ref{Hippo_blur}, \ref{Hippo_rot}, \ref{Amos:mod}, and \ref{kits}, we present qualitative results of various methods under different distribution shifts. As outlined in the main paper, we anticipate a degradation in model performance with increasing shift, accompanied by a corresponding rise in uncertainty. This phenomenon is elaborated upon in our main discussion.
%In Figures \ref{Hippo_blur}, \ref{Hippo_rot}, \ref{Amos:mod} and \ref{kits}, we provide qualitatively results of different methods across different distribution shifts. As mentioned in the main paper, we expect with increasing shift, the performance of the model degrades and this reduction in performance corresponds to high uncertainty.  

\end{document}